\documentclass[10pt,twocolumn,letterpaper]{article}
\usepackage{iccv}
\usepackage{times}
\usepackage{epsfig}
\usepackage{amsmath}
\usepackage{amssymb}
\usepackage{subfigure}
\usepackage{threeparttable}
\usepackage{booktabs}
\newcommand{\myfont}{\fontsize{6.5pt}{\baselineskip}\selectfont}
\newcommand{\tabincell}[2]{\begin{tabular}{@{}#1@{}}#2\end{tabular}}

\iccvfinalcopy 

\ificcvfinal\fi

\begin{document}

\title{LPD-Net: 3D Point Cloud Learning for Large-Scale Place Recognition and Environment Analysis}

\author{Zhe Liu$^1$\thanks{Contact: zheliu@cuhk.edu.hk. The first three authors contributed equally.}, Shunbo Zhou$^1$, Chuanzhe Suo$^1$, Yingtian Liu$^1$, Peng Yin$^3$, Hesheng Wang$^2$, Yun-Hui Liu$^1$\\
\and
$^1$The Chinese University of Hong Kong\\
\and
$^2$Shanghai Jiao Tong University\\
\and
$^3$Carnegie Mellon University\\
}

\maketitle
\ificcvfinal\thispagestyle{empty}\fi

\begin{abstract}
Point cloud based place recognition is still an open issue due to the difficulty in extracting local features from the raw 3D point cloud and generating the global descriptor, and it's even harder in the large-scale dynamic environments.
In this paper, we develop a novel deep neural network, named LPD-Net (Large-scale Place Description Network), which can extract discriminative and generalizable global descriptors from the raw 3D point cloud.
Two modules, the adaptive local feature extraction module and the graph-based neighborhood aggregation module, are proposed, which contribute to extract the local structures and reveal the spatial distribution of local features in the large-scale point cloud, with an end-to-end manner. We implement the proposed global descriptor in solving point cloud based retrieval tasks to achieve the large-scale place recognition. Comparison results show that our LPD-Net is much better than PointNetVLAD and reaches the state-of-the-art. We also compare our LPD-Net with the vision-based solutions to show the robustness of our approach to different weather and light conditions.
\end{abstract}

\section{Introduction}

\begin{figure}[!t]
	\centering
	\includegraphics[width=0.95\columnwidth]{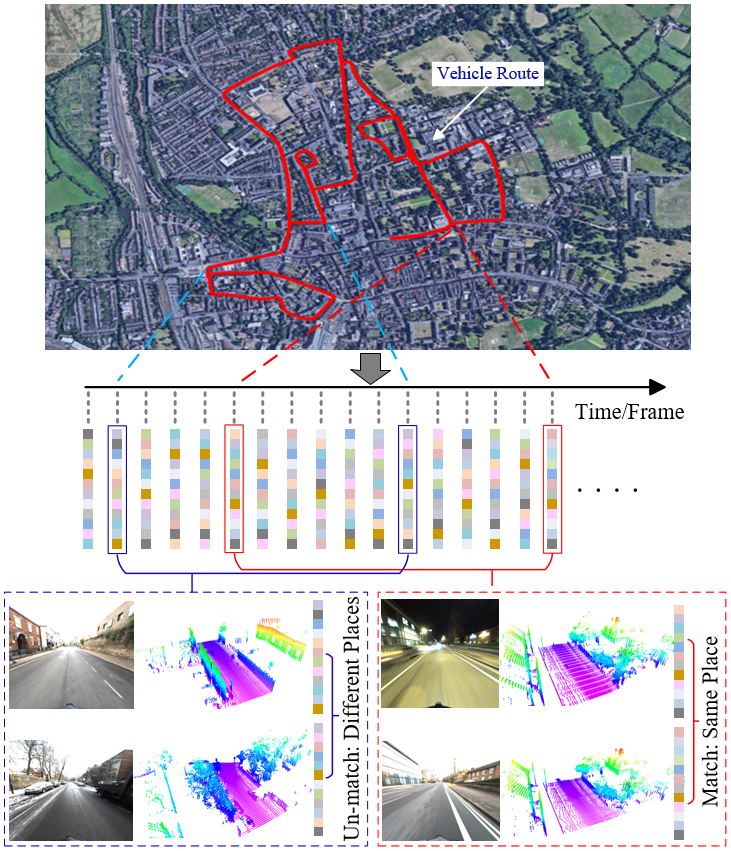}
	\caption{Place recognition in large-scale environments. We use global feature descriptors generated from the raw 3D point cloud data to accomplish place recognition tasks. The lower side shows two examples with different weather and light conditions.}
	\label{figsys}
\end{figure}

Large-scale place recognition is of great importance in robotic applications, such as helping self-driving vehicles to obtain loop-closure candidates, achieve accurate localization and build drift-free globally consistent maps. Vision-based place recognition has been investigated for a long time and lots of successful solutions were presented. Thanks to the feasibility of extracting visual feature descriptors from a query image of a local scene, vision-based approaches have achieved good retrieval performance for place recognitions with respect to the reference map \cite{VisualEA2, VisualPR}. However, vision-based solutions are not robust to season, illumination and viewpoint variations, and also suffer from performance degradations in the place recognition task with bad weather conditions.
Taking into account the above limitations of the vision-based approach, 3D point cloud-based approach provides an alternative option, which is much more robust to season and illumination variations \cite{Pointnetvlad}. By directly using the 3D positions of each point as the network input, PointNet \cite{pointnet} provides a simple and efficient point cloud feature learning framework, but fails to capture fine-grained patterns of the point cloud due to the ignored point local structure information. Inspired by PointNet, different networks have been proposed \cite{pointnet++, dgcnn, kcnet, rwthnet} and achieved advanced point cloud classification and segmentation results with the consideration of well-learned local features. However, 
it is hard to directly implement these networks to extract discriminative and generalizable global descriptors of the point cloud in large scenes. On the other hand, PointNetVLAD \cite{Pointnetvlad} is presented to solve the point cloud description problem in large-scale scenes, but it ignores the spatial distribution of similar local features, which is of great importance in extracting the static structure information in large-scale dynamic environments.

Attempting to address the above issues, we present LPD-Net to extracting discriminative and generalizable global features from large-scale point clouds. As depicted in Fig. \ref{figsys}, based on the generated global descriptor, we solve the point cloud retrieval task for large-scale place recognitions.
Our contributions include:
$1)$ We introduce local features in an adaptive manner as the network input instead of only considering the position of each isolated point, which helps to adequately learn the local structures of the input point cloud. 
$2)$ We propose a graph-based aggregation module in both Feature space and Cartesian space to further reveal the spatial distribution of local features and inductively learn the structure information of the whole point cloud. This contributes to learn a discriminative and generalizable global descriptors for large-scale environments.
$3)$ We utilize the global descriptor for point cloud-based retrieval tasks to achieve large-scale place recognitions. Our LPD-Net outperforms PointNetVLAD in the point cloud based retrieval task and reaches the state-of-the-art. What's more, compared with vision-based solutions, our LPD-Net shows comparable performance and is more robust to different weather and light conditions.

\section{Related Work}

Handcrafted local features, such as the histogram feature \cite{Handfpfh}, the inner-distance-based descriptor \cite{Handinnerdis} and the heat kernel signatures \cite{Handheat}, are widely used for point cloud-based recognition tasks, but they are usually designed for specific applications and have a poor generalization ability.
In order to solve these problems, deep learning based methods was presented for point cloud feature descriptions.
Convolution neural network (CNN) has achieved amazing feature learning results for regular 2D image data. However, it is hard to extend the current CNN-based method to 3D point clouds due to their orderless.
Some researches attempt to solve this problem by describing the raw point cloud by a regular 3D volume representation, such as the 3D ShapeNets \cite{ModelNet}, VoxNet \cite{VoxNet}, volumetric CNNs \cite{Volumetriccnn}, VoxelNet \cite{VoxelNet} and 3D-GAN \cite{3DGAN}. 
Some other methods, such as the DeepPano \cite{DeepPano} and Multiview CNNs \cite{Multiviewcnns}, project 3D point clouds into 2D images and use the 2D CNN to learn features. However, these approaches usually introduce quantization errors and high computational cost, hence hard to capture high-resolution features with high update rate.

PointNet \cite{pointnet} achieves the feature learning directly from the raw 3D point cloud data for the first time. 
As an enhanced version, PointNet++ \cite{pointnet++} introduces the hierarchical feature learning to learn local features with increasing scales, but it still only operates each point independently during the learning process. Ignoring the relationship between local points leads to the limitation ability of revealing local structures of the input point cloud. To solve this, DG-CNN \cite{dgcnn} and KC-Net \cite{kcnet} mine the neighborhood relations through the dynamic graph network and the kernel correlation respectively. Moreover, \cite{rwthnet} captures local features by performing the kNN algorithm in the feature space and the k-means algorithm in the initial word space simultaneously.
However, they obtain fine-grained features at the expense of ignoring the feature distribution information. What's more, the performance of these approaches in large-scale place recognition tasks have not been validated. 

Traditional point cloud-based large-scale place recognition algorithms \cite{xutransIIB} usually rely on a global, off-line, and high-resolution map, and can achieve centimeter-level localization, but at the cost of time-consuming off-line map registration and data storage requirements. SegMatch \cite{Segmatch} presents a place matching method based on local segment descriptions, but they need to build a dense local map by accumulating a stream of original point clouds to solve the local sparsity problem. PointNetVLAD \cite{Pointnetvlad} achieves the state-of-the-art place recognition results. 
However, as mentioned before, it does not consider the local structure information and ignores the spatial distribution of local features. These factors, however, is proved in our ablation studies that will greatly improve the place recognition results. 

\begin{figure*}[!t]
	\centering
	\includegraphics[width=2.1\columnwidth]{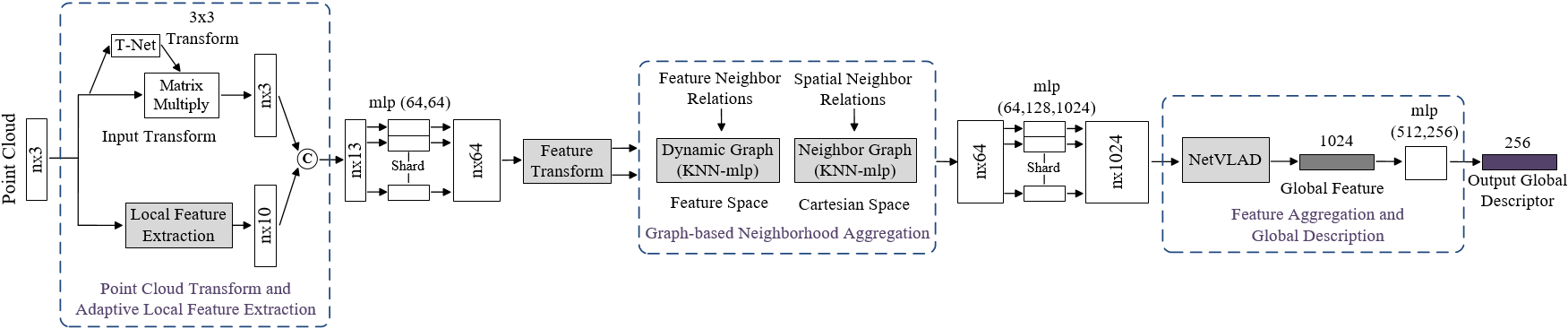}
	\caption{\textbf{LPD-Net Architecture.} The network takes the raw point cloud data as input, applies Adaptive Local Feature Extraction to obtain the point cloud distribution and the enhanced local features, which are aggregated both in the Feature Space and the Cartesian Space through the graph neural network. The resulted feature vectors are then utilized by NetVLAD \cite{netvlad} to generate a global descriptor.}
	\label{fignet}
	\vspace{-0.3cm}
\end{figure*}

\section{Network Design}
The objective of our LPD-Net is to extract discriminative and generalizable global descriptors from the raw 3D point cloud directly, and based on which, to solve the point cloud retrieval problems. Using the extracted global descriptor, the computational and storage complexity will be greatly reduced, thus enabling the real-time place recognition tasks. We believe that the obtained place recognition results will greatly facilitate the loop closure detection, localization and mapping tasks in robotics and self-driving applications.

\subsection{The Network Architecture}
As we mentioned above, most of the existing work is done on the small-scale object point cloud data (e.g. ModelNet \cite{ModelNet} and ShapeNet \cite{ShapeNet}), but this is not the case for large-scale environments, since such point clouds are mainly composed of different objects in the scene and with unknown relationships between the objects. In contrast, we have customized for large-scale environments and proposed a network with three main modules, $1)$ Feature Network (FN), $2)$ Graph-based Neighborhood Aggregation, and $3)$ NetVLAD \cite{netvlad}. The complete network architecture of LPD-Net is shown in Fig. \ref{fignet}. The NetVLAD is designed to aggregate local feature descriptors and generate the global descriptor vector for the input data. Similar to \cite{Pointnetvlad}, the loss function of the network uses lazy quadruplet loss based on metric learning, so that the positive sample distance is reduced during the training process and the negative sample distance is enlarged to obtain a unique scene description vector. In addition, it has been proven to be permutation invariant, thus suitable for 3D point cloud. 

\subsection{Feature Network}

Existing networks \cite{pointnet,pointnet++,Pointnetvlad} only use the point position as the network input, local structures and point distributions have not been considered. This limits the feature learning ability \cite{sonet}. Local features usually represent the generalized information in the local neighborhood of each point, and it has been successfully applied to different scene interpretation applications \cite{Handfeature,Segmatch}.
Inspired by this, 
our FN introduces local features to capture the local structure around each point.

\subsubsection{Feature Network Structure}

The raw point cloud data is simultaneously input to the Input Transformation Net \cite{pointnet} and the Adaptive Local Feature Extractor (as will be introduced in Section \ref{seclocalfeature}), the former aims to ensure the rotational translation invariance \cite{pointnet} of the input point coordinates, and the latter aims to fully consider the statistical local distribution characteristics. 
It should be noted that, the point cloud acquired in large-scale scenes often has uneven local point distributions, which may affect the network accuracy. To handle this, the adaptive neighborhood structure is considered to select the appropriate neighborhood size according to different situations to fuse the neighborhood information of each point.
We then map the above two kinds of features (with the concatenation operation) to the high-dimensional space, and finally make the output of FN invariant to the spatial transformation through the Feature Transformation Net.

\begin{figure*}[!t]
	\centering
	\subfigure[FN-Original structure (O)]{\includegraphics[width=0.5\columnwidth]{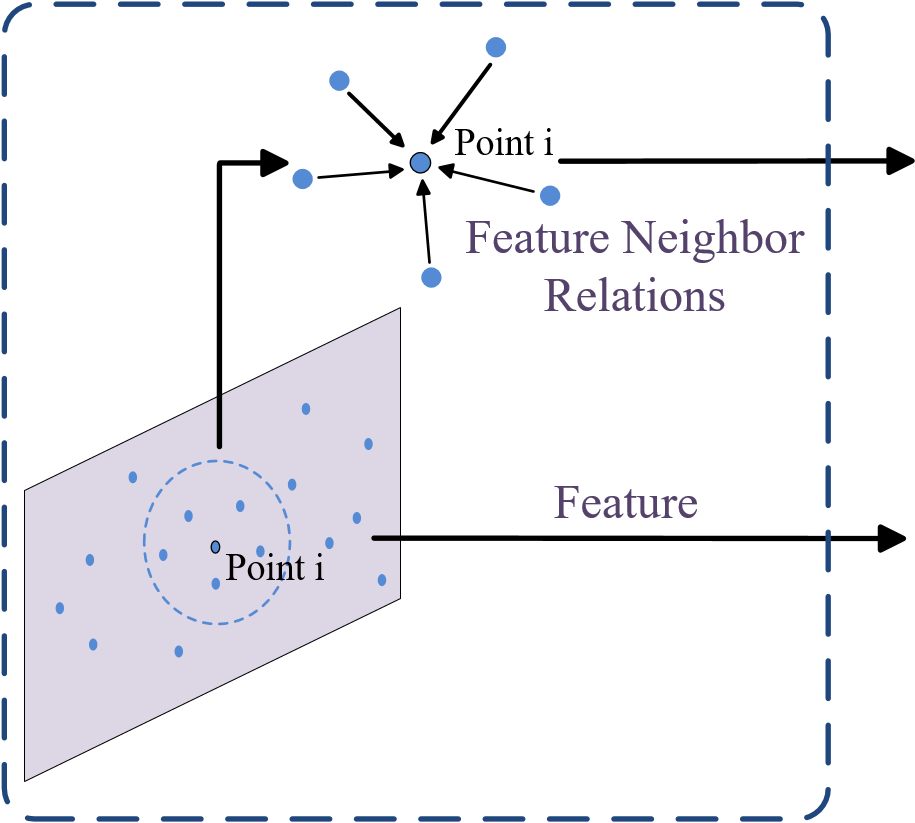} \label{fig:fnr:1}}
	\subfigure[FN-Series structure (S)]{\includegraphics[width=0.75\columnwidth]{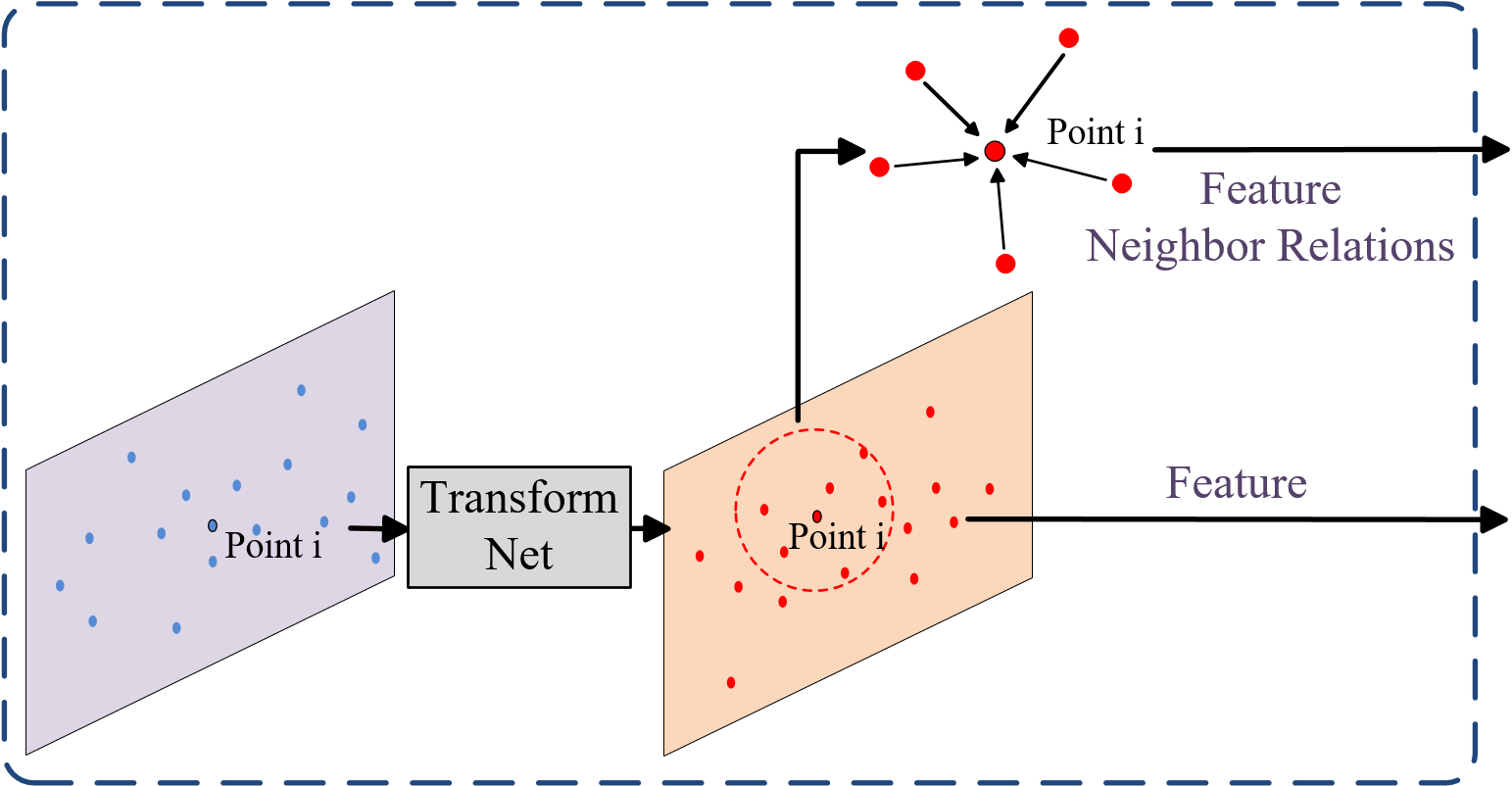} \label{fig:fnr:2}}
	\subfigure[FN-Parallel structure (P)]{\includegraphics[width=0.6\columnwidth]{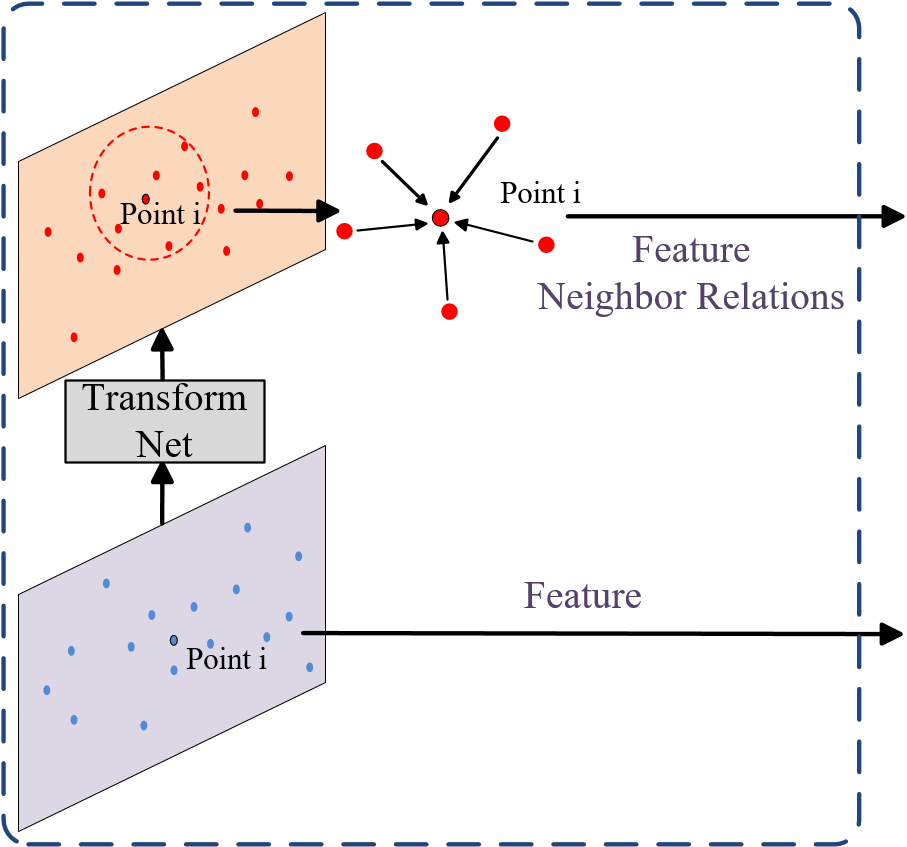} \label{fig:fnr:3}}
	\caption{Different network structures for feature transformation and relation extraction.}
	\label{fig:fnr}
	\vspace{-0.3cm}
\end{figure*}

\subsubsection{Adaptive Local Feature Extraction}\label{seclocalfeature}

We introduce local distribution features by considering the local 3D structure around each point $i$. $k$ nearest neighboring points are counted and the respective local 3D position covariance matrix is considered as the local structure tensor. Without loss of generality, we assume that $\lambda^i_1\ge\lambda^i_2\ge\lambda^i_3\ge0$ represent the eigenvalues of the symmetric positive-define covariance matrix. According to \cite{Handfeature}, the following measurement can be used to describe the unpredictability of the local structure from the aspect of the Shannon information entropy theory,
\begin{equation}
E_{i}=-L_i\ln{L_i}-P_i\ln{P_i}-S_i\ln{S_i},
\end{equation}
where $L_i=\frac{\lambda^i_1-\lambda^i_2}{\lambda^i_1}$, $P_i=\frac{\lambda^i_2-\lambda^i_3}{\lambda^i_1}$ and $S_i=\frac{\lambda^i_3}{\lambda^i_1}$ represent the linearity, planarity and scattering features of the local neighborhood of each point respectively. These features describe the 1D, 2D and 3D local structures around each point \cite{Handfeature}. Since the point distribution in a point cloud is typically uniform, we adaptively choose the neighborhood of each point $i$ by minimizing $E_i$ across different $k$ values and the optimal neighbor size is determined as
\begin{equation}
k^{i}_{opt}=\arg\min_{k} E_i(k).
\end{equation}

Local features suitable for describing large-scale scenes can be classified into four classes: eigenvalue-based 3D features ($F_{3D}$), features arising from the projection of the 3D point onto the horizontal plane ($F_{2D}$), normal vector-based features ($F_V$), and features based on Z-axis statistics ($F_Z$). Existing researches have validated that $F_{3D}$, $F_V$ and $F_Z$ are effective in solving the large-scale 3D scene analysis problem \cite{Handfeature}, and $F_{2D}$ and $F_Z$ are effective in solving the large-scale localization problem in self-driving tasks\cite{baiducvpr,Segmatch}. Considering the feature redundancy and discriminability, we select the following ten local features to describe the local distribution and structure information around each point $i$:
\begin{itemize}
\item $F_{3D}$ features: Change of curvature $C_i=\frac{\lambda^i_3}{\sum_{j=1}^3 \lambda^i_j}$, Omni-variance $O_i=\frac{\sqrt[3]{\prod_{j=1}^3 \lambda^i_j}}{\sum_{j=1}^3 \lambda^i_j}$, Linearity $L_i=\frac{\lambda^i_1-\lambda^i_2}{\lambda^i_1}$, Eigenvalue-entropy $A_i=-\sum_{j=1}^3(\lambda^i_j\ln\lambda^i_j)$, and Local point density $D_i=\frac{k^i_{opt}}{\frac{4}{3}\prod_{j=1}^3 \lambda^i_j}$.
\item $F_{2D}$ features: 2D scattering $S_{i,2D}=\lambda^i_{2D,1}+\lambda^i_{2D,2}$ and 2D linearity $L_{i,2D}=\frac{\lambda^i_{2D,2}}{\lambda^i_{2D,1}}$, where $\lambda^i_{2D,1}$ and $\lambda^i_{2D,2}$ represent the eigenvalues of the corresponding 2D covariance matrix.
\item $F_V$ feature: Vertical component of normal vector $V_i$.
\item $F_Z$ features: Maximum height difference $\Delta Z_{i,\max}$ and Height variance $\sigma Z_{i,var}$.
\end{itemize}

\subsubsection{Feature Transformation and Relation Extraction}

In the output of the Adaptive Local Feature Extraction module, each data can be regarded as the feature description of the surrounding neighborhood since we have merged the neighborhood structure into the feature vector of the neighborhood center point.
Three structures are then designed in the Feature Transform module shown in Fig. \ref{fignet} to further reveal the relations between the local features:
\begin{itemize}
\item FN-Original structure (Fig. \ref{fig:fnr:1}): The two outputs are the feature vector $f_{F}$ and the neighborhood relation vector $f_{R}$ by performing kNN operations on $f_{F}$.
\item FN-Series structure (Fig. \ref{fig:fnr:2}): The two outputs are the feature vector $f_{FT}$ which has been transformed by the Transform Net \cite{pointnet}, and the neighborhood relation vector $f_{RT}$ by performing kNN operations on $f_{FT}$.
\item FN-Parallel structure (Fig. \ref{fig:fnr:3}): The two outputs are the feature vector $f_{F}$ and the neighborhood relation vector $f_{RT}$, where $f_{RT}$ is the same with that in FN-Series structure. 
\end{itemize}
The ablation study in Section \ref{secablation} reveals that the FN-Parallel structure is the best one in our case.

\begin{figure}[!h]
	\centering
	\includegraphics[width=1\columnwidth]{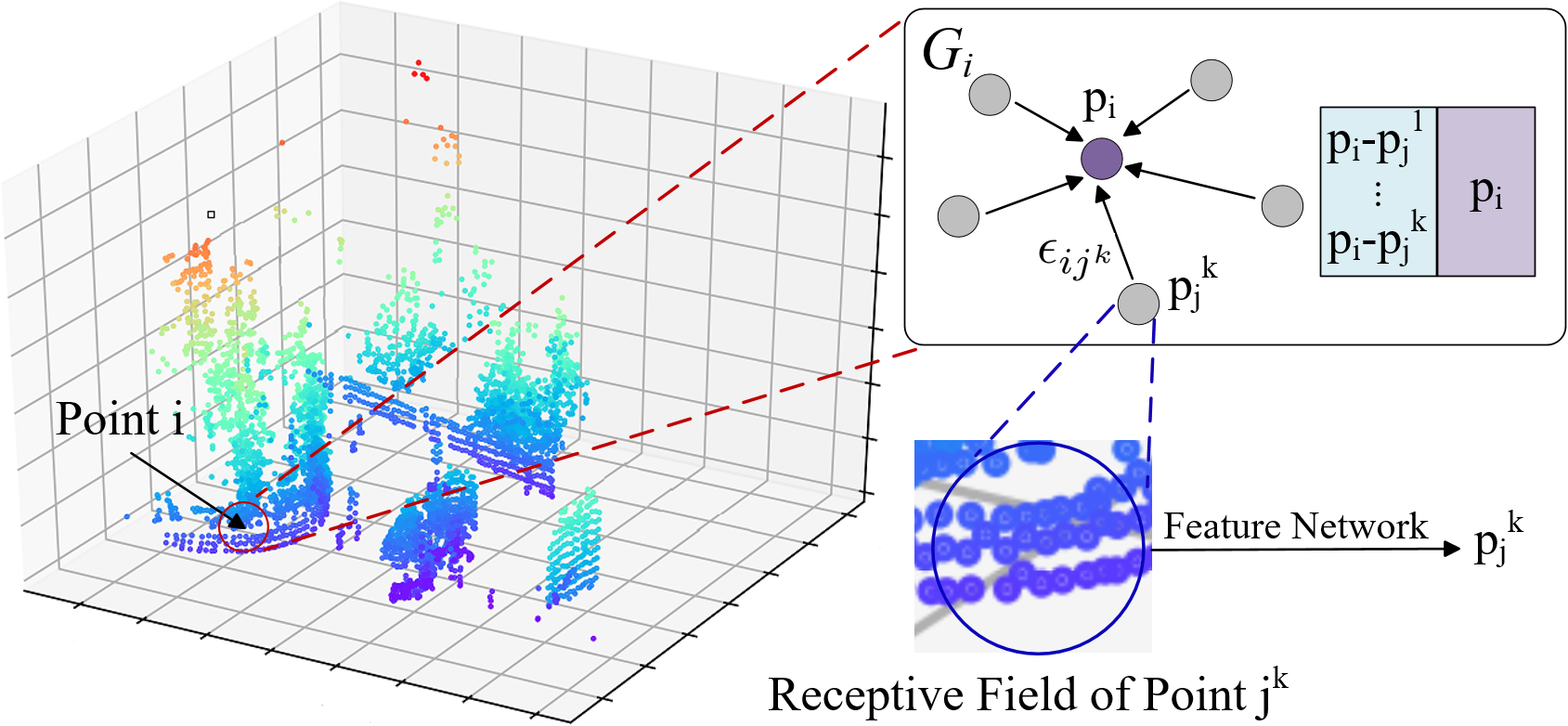}
	\caption{Graph formulation. Note that the receptive field of each point corresponds to a local neighborhood in the original point cloud, since the FN has introduced the local structure into the feature of each point. Then we utilize GNN for feature aggregations.}
	\label{figfield}
\end{figure}

\begin{figure}[!h]
	\centering
	\includegraphics[width=1\columnwidth]{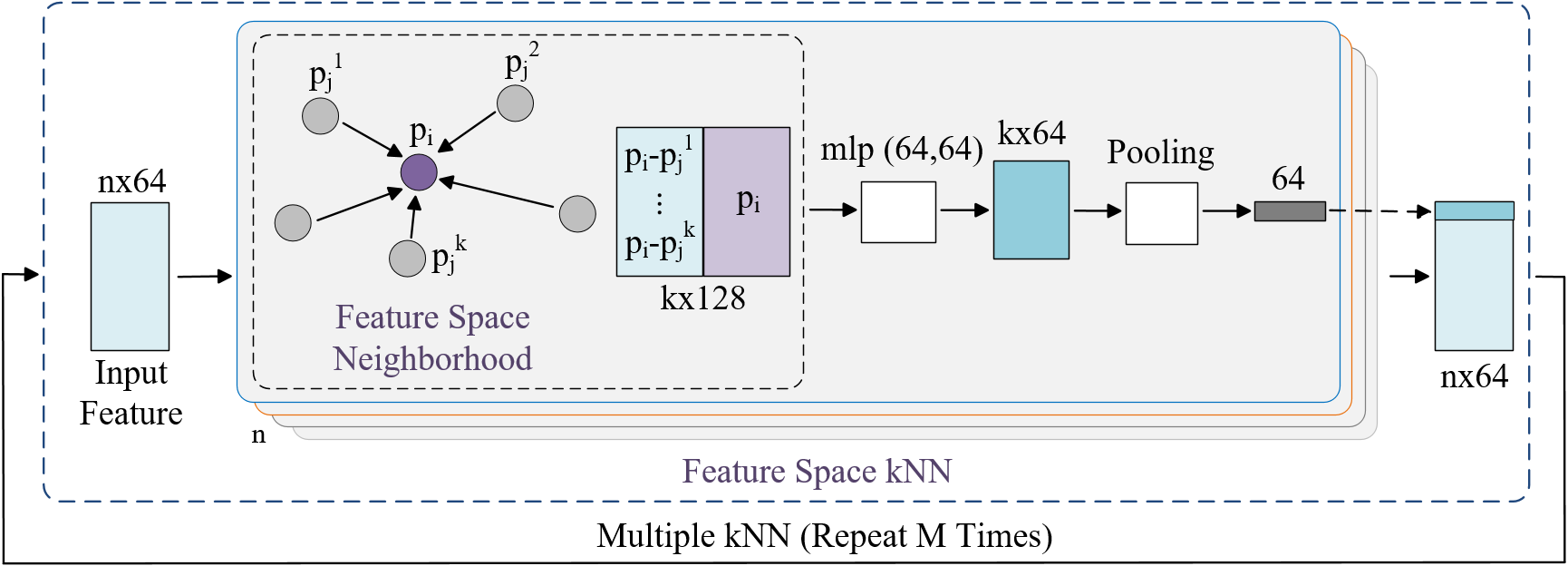}
	\caption{Feature space graph-based neighborhood aggregation.}
	\label{figknn1}
	\vspace{-0.2cm}
\end{figure}

\begin{figure*}[!t]
	\centering
	\subfigure[Prarllel-Concatenation structure (PC)]{\includegraphics[width=0.6\columnwidth]{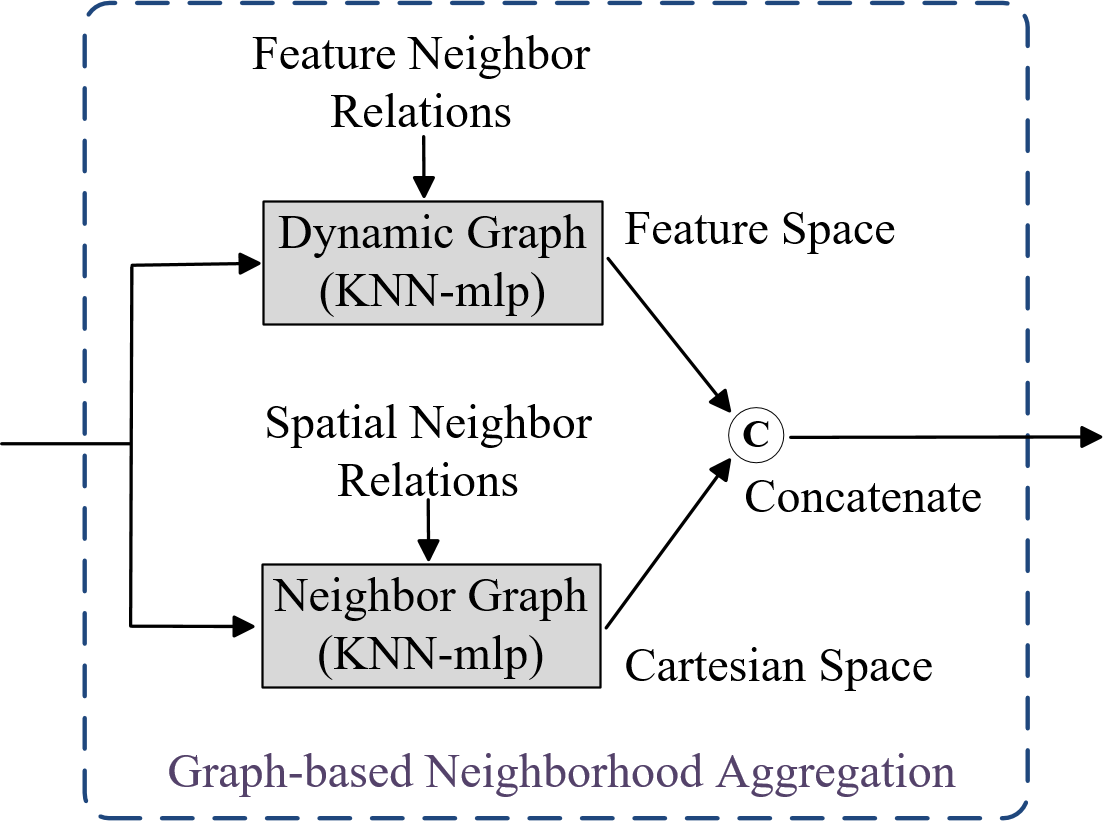} \label{figs:1}}
	\subfigure[Parallel-Maxpooling structure (PM)]{\includegraphics[width=0.6\columnwidth]{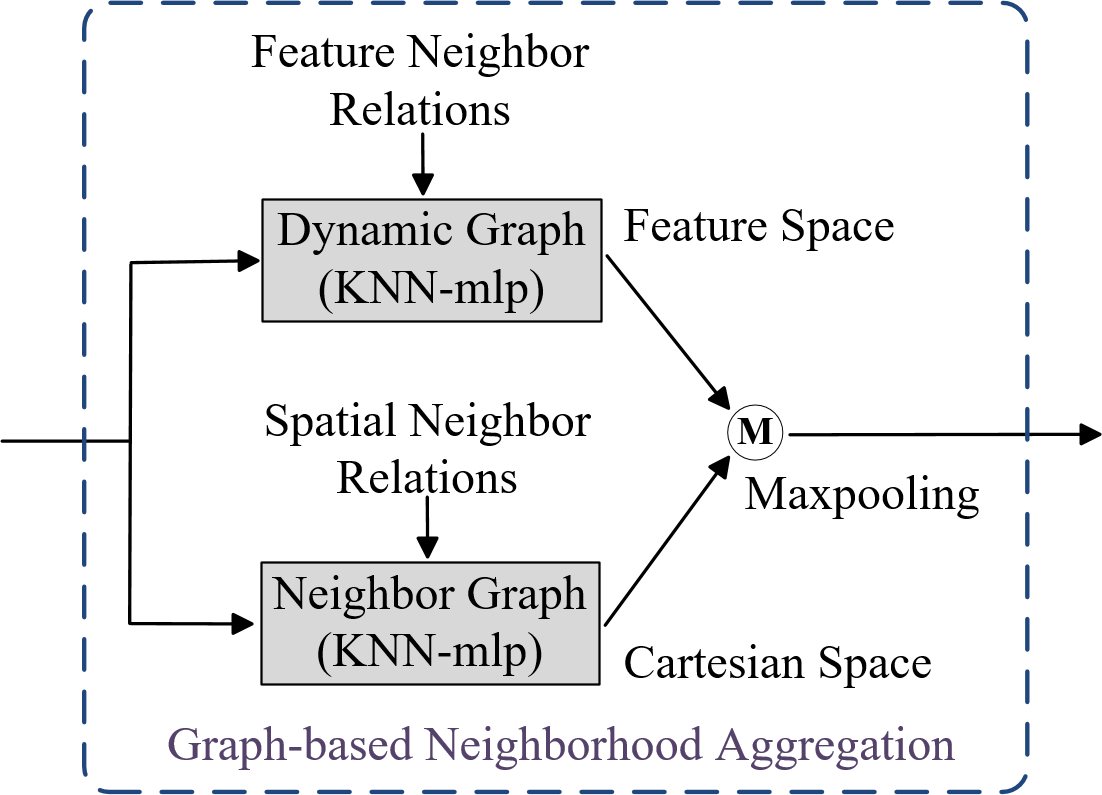} \label{figs:2}}
	\subfigure[Series-FC structure (SF)]{\includegraphics[width=0.55\columnwidth]{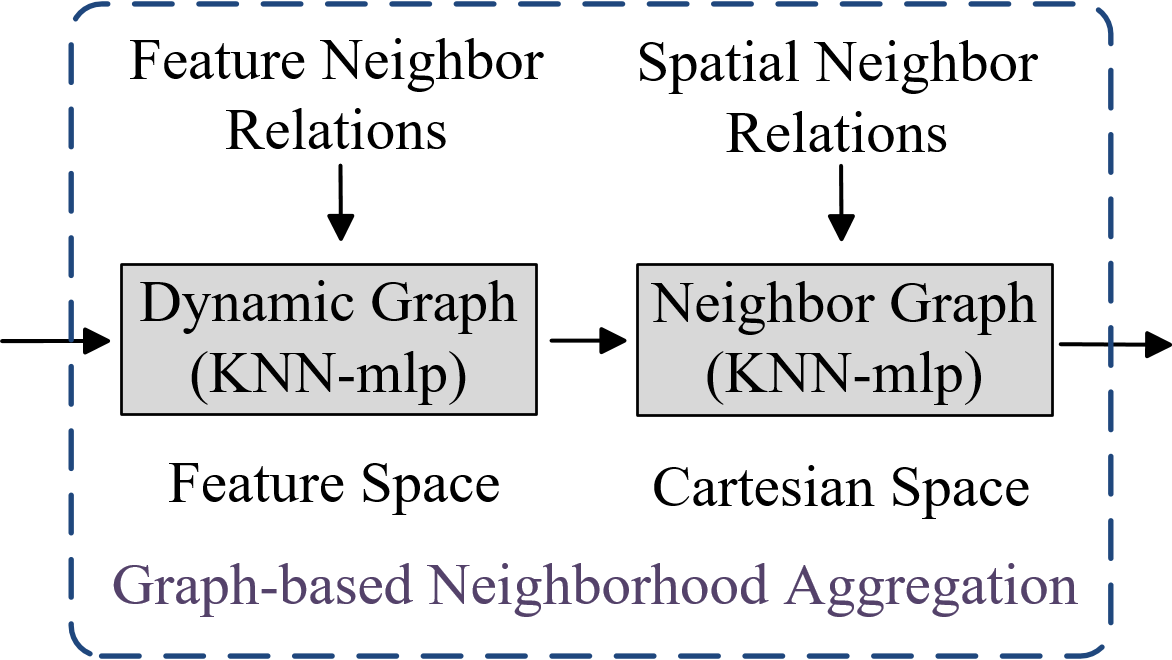} \label{figs:3}}
	\caption{Different network structures for feature aggregation.}
	\label{figs}
	\vspace{-0.3cm}
\end{figure*}

\subsection{Graph-based Neighborhood Aggregation}

Different with the object point clouds, the point clouds of large-scale environments mostly contain several local 3D structures (such as planes, corners, shapes, etc.) of surrounding objects.
Similar local 3D structures which locate in different parts of the point cloud usually have similar local features. Their spatial distribution relationships are also of great importance in place description and recognition tasks. We introduce the relational representation from the Graph Neural Network (GNN) \cite{gnn} into our LPD-Net, which uses a structured representation to get the compositions and their relationship. Specifically, we represent the compositions of the scene as the nodes in the graph model (Fig. \ref{figfield}), and represent their intrinsic relationships and generate unique scene descriptors through GNN.

\subsubsection{Graph Neural Network Structure}

The outputs of the Feature Network (the feature vector and the neighborhood relation vector) are used as the input of the graph network, and feature aggregation is performed in both the feature space and the Cartesian space.
As shown in Fig. \ref{figknn1}, in the feature space, we build a dynamic graph $G_{i,d}$ for each point $i$ through the multiple kNN iterations.
More specifically, in each iteration, the output feature vector of the previous iteration is used as the network input and a kNN aggregation is conducted on each point by finding $k$ neighbors with the nearest feature space distances. This is similar to CNN to achieve the multi-scale feature learning.
Each point feature $p_i$ is treated as a node in the graph. Each edge $\epsilon_{ij}^m$ represents the feature space relation between $p_i$ and its $k$ nearest neighbors $p_j^m$ in the feature space, and $\epsilon_{ij}^m$ is defined as $\epsilon_{ij}^m=p_i-p_j^m, m={1, 2, ..., k}$. The mlp network is used to update neighbor relations and the max pooling operation is used to aggregate $k$ edge information into a feature vector to update the point feature $p_i$.
Note that the features of two points with a large Cartesian space distance can also be aggregated for capturing similar semantic information, due to the graph-based feature learning in the feature space.
In addition, the neighborhood information in the Cartesian space should also be concerned. The kNN-graph network is also implemented in the Cartesian space. The node and edge are defined as the same in the feature space and the only difference is that we consider the Euclidean distance to build the kNN relations.

\subsubsection{Feature Aggregation Structure}

In LPD-Net, GNN modules in the feature space and the Cartesian space aggregate neighborhood features and spatial distribution information separately. 
We designed three different structures to further aggregate these two modules:
\begin{itemize}
\item Prarllel-Concatenation structure (PC, Fig. \ref{figs:1}): Cascade the output feature vectors of the two modules and merge the dual-dimensional information through MLP to aggregate the features.
\item Parallel-Maxpooling structure (PM, Fig. \ref{figs:2}): Directly integrate the output feature vectors of the two models through the max pooling layer, taking the maximum values to generate the unified feature vector. 
\item Series-FC structure (SF, Fig. \ref{figs:3}): The output feature vector of one module is utilized as the input feature of the other module. 
\end{itemize}
The experimental result in Section \ref{secmainresult} reveals that the SF structure with the order shown in Fig. \ref{figs:3} is the best one in our case. 

\subsection{Discussion}

Based on the proposed LPD-Net, we can analyze the environment by studying the statistical characteristics of all the global descriptors, such as calculating the similarity of two places by the $L_2$ distance between the two corresponding global descriptors, or evaluating the uniqueness of each place by calculating its distance to all the other places. More details can be found in our supplementary materials. 

\section{Experiments}

\begin{table*}[!t]
\tiny
\renewcommand{\arraystretch}{1.0}
\caption{LDP-Net configuration.}
\label{tabconf}
\centering
\begin{tabular}{|c|c|c|c|c|c|c|c|c|c|c|c|c|c|}
\hline
\multicolumn{2}{|c|}{NN-VLAD} & \multicolumn{2}{|c|}{FN-VLAD} & \multicolumn{2}{|c|}{FN-NG-VLAD} & \multicolumn{2}{|c|}{FN-DG-VLAD} & \multicolumn{2}{|c|}{FN-PM-VLAD} & \multicolumn{2}{|c|}{FN-PC-VLAD} & \multicolumn{2}{|c|}{FN-SF-VLAD}\\
\hline
\hline
point-3 & mlp-10 & point-3 & ALF-10 & point-3 & ALF-10 & point-3 & ALF-10 & point-3 & ALF-10 & point-3 & ALF-10 & point-3 & ALF-10\\
\hline
T-Net-3 & & T-Net-3 & & T-Net-3 & & T-Net-3 & & T-Net-3 & & T-Net-3 & & T-Net-3 & \\
\hline
\multicolumn{2}{|c|}{concat-13} & \multicolumn{2}{|c|}{concat-13} & \multicolumn{2}{|c|}{concat-13} & \multicolumn{2}{|c|}{concat-13} & \multicolumn{2}{|c|}{concat-13} & \multicolumn{2}{|c|}{concat-13} & \multicolumn{2}{|c|}{concat-13}\\
\hline
\multicolumn{2}{|c|}{mlp-64} & \multicolumn{2}{|c|}{mlp-64} & \multicolumn{2}{|c|}{mlp-64} & \multicolumn{2}{|c|}{mlp-64} & \multicolumn{2}{|c|}{mlp-64} & \multicolumn{2}{|c|}{mlp-64} & \multicolumn{2}{|c|}{mlp-64}\\
\multicolumn{2}{|c|}{mlp-64} & \multicolumn{2}{|c|}{mlp-64} & \multicolumn{2}{|c|}{mlp-64} & \multicolumn{2}{|c|}{mlp-64} & \multicolumn{2}{|c|}{mlp-64} & \multicolumn{2}{|c|}{mlp-64} & \multicolumn{2}{|c|}{mlp-64}\\
\hline
\hline
\multicolumn{14}{|c|}{Feature transform-64 and relation extraction-feature space KNN (Kf) \& Cartesian space KNN (Kc) }\\
\hline
\multicolumn{2}{|c|}{} & \multicolumn{2}{|c|}{} & \multicolumn{2}{|c|}{KNN-Kc*64} & \multicolumn{2}{|c|}{KNN-Kf*64} & KNN-Kf*64 & KNN-Kc*64 & KNN-Kf*64 & KNN-Kc*64 & \multicolumn{2}{|c|}{KNN-Kf*64}\\
\multicolumn{2}{|c|}{} & \multicolumn{2}{|c|}{} & \multicolumn{2}{|c|}{mlp-64} & \multicolumn{2}{|c|}{EF-k*128} & EF-k*128 & mlp-64 & EF-k*128 & mlp-64 & \multicolumn{2}{|c|}{EF-k*128}\\
\multicolumn{2}{|c|}{} & \multicolumn{2}{|c|}{} & \multicolumn{2}{|c|}{mlp-64} & \multicolumn{2}{|c|}{mlp-64} & mlp-64 & mlp-64 & mlp-64 & mlp-64 & \multicolumn{2}{|c|}{mlp-64}\\
\multicolumn{2}{|c|}{} & \multicolumn{2}{|c|}{} & \multicolumn{2}{|c|}{} & \multicolumn{2}{|c|}{mlp-64} & mlp-64 & & mlp-64 & & \multicolumn{2}{|c|}{mlp-64}\\
\hline
\multicolumn{2}{|c|}{} & \multicolumn{2}{|c|}{} & \multicolumn{2}{|c|}{maxpooling-64} & \multicolumn{2}{|c|}{maxpooling-64} & \multicolumn{2}{|c|}{maxpooling-64} &\multicolumn{2}{|c|}{concat-64} & \multicolumn{2}{|c|}{maxpooling-64}\\
\hline
\multicolumn{2}{|c|}{} & \multicolumn{2}{|c|}{} & \multicolumn{2}{|c|}{} & \multicolumn{2}{|c|}{} & \multicolumn{2}{|c|}{} &\multicolumn{2}{|c|}{} & \multicolumn{2}{|c|}{KNN-Kc*64}\\
\multicolumn{2}{|c|}{} & \multicolumn{2}{|c|}{} & \multicolumn{2}{|c|}{} & \multicolumn{2}{|c|}{} & \multicolumn{2}{|c|}{} &\multicolumn{2}{|c|}{} & \multicolumn{2}{|c|}{mlp-64}\\
\multicolumn{2}{|c|}{} & \multicolumn{2}{|c|}{} & \multicolumn{2}{|c|}{} & \multicolumn{2}{|c|}{} & \multicolumn{2}{|c|}{} &\multicolumn{2}{|c|}{} & \multicolumn{2}{|c|}{mlp-64}\\
\multicolumn{2}{|c|}{} & \multicolumn{2}{|c|}{} & \multicolumn{2}{|c|}{} & \multicolumn{2}{|c|}{} & \multicolumn{2}{|c|}{} &\multicolumn{2}{|c|}{} & \multicolumn{2}{|c|}{maxpooling-64}\\
\hline
\hline
\multicolumn{14}{|c|}{FC-64}\\
\hline
\multicolumn{14}{|c|}{FC-128}\\
\hline
\multicolumn{14}{|c|}{FC-1024}\\
\hline
\multicolumn{14}{|c|}{L2-normalization}\\
\hline
\multicolumn{14}{|c|}{NetVLAD-D}\\
\hline
\multicolumn{14}{|c|}{L2-normalization}\\
\hline
\multicolumn{14}{|c|}{Lazy Quadruplet Loss}\\
\hline
\end{tabular}
\begin{tablenotes}
\item[1] ALF: Adaptive local feature.
\end{tablenotes}
\vspace{-0.3cm}
\end{table*}

The configuration of LPD-Net is shown in Tab. \ref{tabconf}. In NetVLAD \cite{netvlad,Pointnetvlad}, the lazy quadruplet loss paremeters are set as $\alpha=0.5, \beta=0.2, P_{pos}=2, P_{neg}=18$. We train and evaluate the network on the modified Oxford Robotcar dataset presented by \cite{Pointnetvlad}, which includes 44 data sets from the original Robotcar dataset, with 21,711 training submaps and 3030 testing submaps. 
We also directly transplant the trained model to the In-house Dataset \cite{Pointnetvlad} 
for evaluation and verify its generalization ability.
Please not that in all datasets, the point data has been randomly down-sampled to 4096 points and normalized to [-1,1]. More details of the datasets can be found in \cite{Pointnetvlad}.
All experiments are conducted with a 1080Ti GPU on TensorFlow.

\subsection{Place Recognition Results}\label{secmainresult}

\begin{figure}[!h]
\centering
\includegraphics[width=0.8\columnwidth]{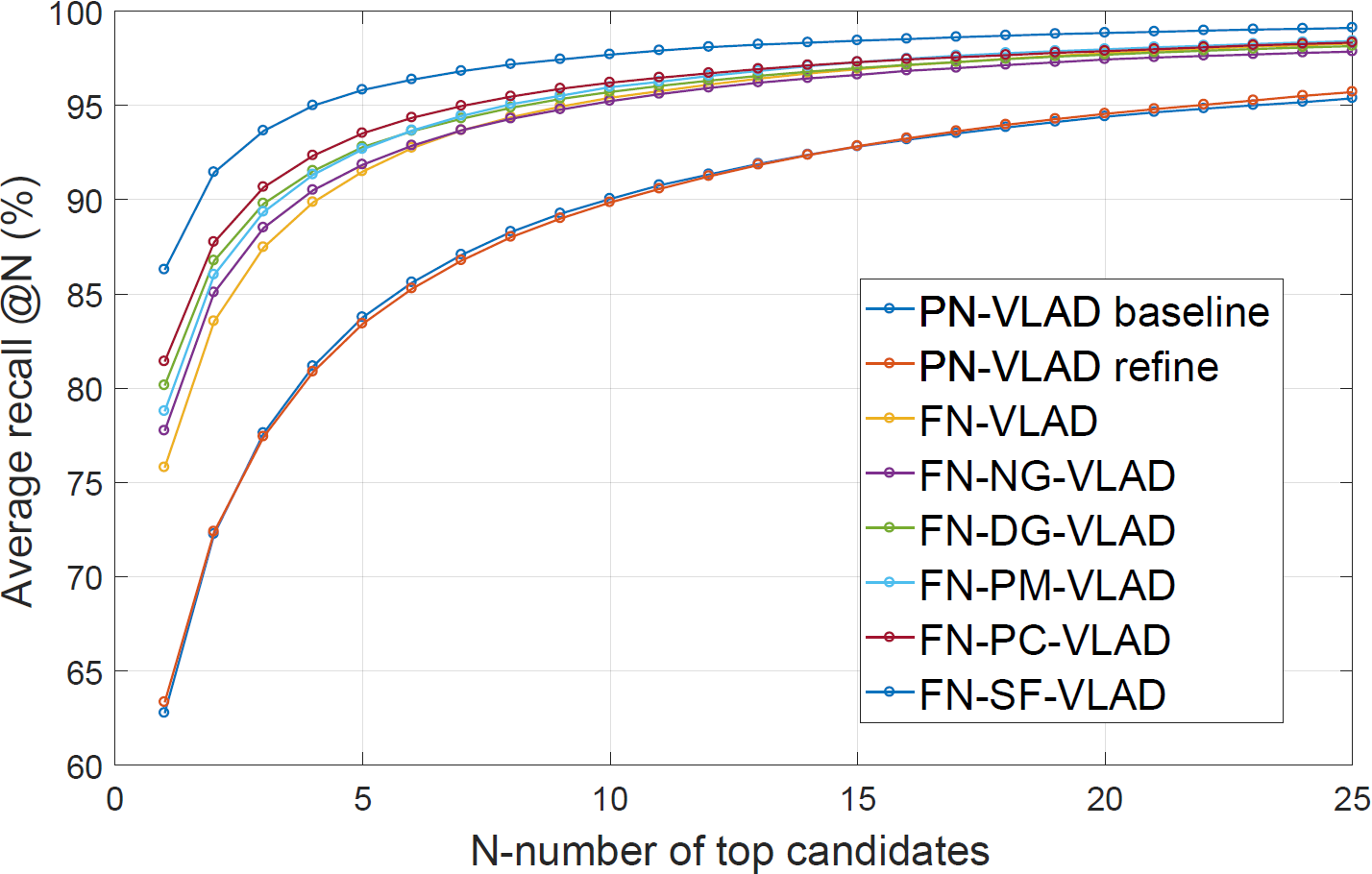}
\caption{Average recall under different networks.}
\label{figexp1}
\end{figure}

\begin{table}[!h]
\small
\renewcommand{\arraystretch}{1.0}
\caption{Comparison results of the average recall (\%) at top 1\% (@1\%) and at top 1 (@1) under different networks.}
\label{tabexp1}
\centering
\begin{tabular}{c|c|c}
\hline
\hline
 & Ave recall @1\% & Ave recall @1\\
\hline
PN STD & $46.52$ & $31.87$\\
PN MAX & $73.87$ & $54.16$\\
PN-VLAD baseline$^*$ & $81.01$ & $62.76$\\
PN-VLAD refine$^*$ & $80.71$ & $63.33$\\
\hline
NN-VLAD (our) & $79.21$ & $61.96$\\
FN-VLAD (our) & $89.77$ & $75.79$\\
FN-NG-VLAD (our) & $90.38$ & $77.74$\\
FN-DG-VLAD (our) & $91.44$ & $80.14$\\
FN-PM-VLAD (our) & $91.20$ & $78.77$\\
FN-PC-VLAD (our) & $92.27$ & $81.41$\\
FN-SF-VLAD (our) & $\bf{94.92}$ & $\bf{86.28}$\\
\hline
\hline
\end{tabular}
\begin{tablenotes}
\item $^*$This result is obtained by using their open-source programs.
\end{tablenotes}
\end{table}

\begin{table}[!h]
\footnotesize
	\renewcommand{\arraystretch}{1.0}
\caption{Comparison results of the memory and computation required under different networks.}
	\centering
	\begin{tabular}{cccc}
		\hline
		\hline
		& Parameters & FLOPs & Runtime per frame\\
		\hline
        PN-VLAD baseline & 1.978M & 411M & 13.09ms\\
		FN-PM-VLAD (our) & 1.981M & 749M & 29.23ms\\
		FN-PC-VLAD (our) & 1.981M & 753M & 27.03ms\\
        FN-SF-VLAD (our) & 1.981M & 749M & 23.58ms\\		
		\hline
		\hline
	\end{tabular}
	\label{tabexptime}
\begin{tablenotes}
\item FLOPs: required floting-point operations.
\end{tablenotes}
	\vspace{-0.4cm}
\end{table}

The selected Robotcar dataset contains the point clouds collected in various season and weather conditions and different times. We query the same scene in these different sets for place recognition tasks. 
Specifically, we use the LPD-Net to generate the global descriptors and query the scene with the closest $L_2$ distance (in the descriptor space) to the test scene to determine whether it is the same place.
Similar to \cite{Pointnetvlad}, the Recall indices, including the Average Recall@N and Average Recall@1\%, are utilized to evaluate the place recognition accuracy. 
We compare our LPD-Net with the original PointNet architecture with the maxpool layer (PN MAX) and the PointNet trained for object classification in ModelNet (PN STD) to see whether the model trained on small-scale object datesets can be scaled to large-scale cases. We also compare our LPD-Net with the state-of-the-art PN-VLAD baseline and PN-VLAD refine \cite{Pointnetvlad}. We evaluate the PN STD, PN MAX, PN-VLAD baseline and PN-VLAD refine on the Oxford training dataset. The network configurations of PN STD, PN MAX, PN-VLAD baseline and refine are set to be the same as \cite{pointnet,Pointnetvlad}.

Comparison results are shown in Fig. \ref{figexp1} and Tab. \ref{tabexp1}, where FN-PM-VLAD, FN-PC-VLAD, and FN-SF-VLAD represent our network with the three different feature aggregation structures PM, PC, and SF. FN-VLAD is our network without the graph-based neighborhood aggregation module. DG and NG represent the Dynamic Graph and Neighbor Graph in the proposed graph-based neighborhood aggregation module. Additionally, we also design the NeuralNeighborVLAD network (NN-VLAD), which uses kNN clustering (k=20) and mlp module to replace the adaptive local feature extraction module presented in Section \ref{seclocalfeature}. The output of the network is also a 10 dimensional neighborhood feature, and the features are obtained through network learning.
Thanks to the adaptive local feature extraction and graph neural network modules, our LPD-Net has superior advantages for place recognition in large-scale environments.
What's more, among the three aggregation structures, FN-SF-VLAD is the best one, far exceeding PointNetVLAD from $81.01\%$ to $94.92\%$ at top 1\% (unless otherwise stated, the LPD-Net represents the FN-SF-VLAD in this paper). In SF, the graph neural network learns the neighborhood structure features of the same semantic information in the feature space, and then further aggregates them in the Cartesian space. So we believe that SF can learn the spatial distribution characteristics of neighborhood features, which is of great importance for large-scale place recognitions. 
In addition, PC is better than PM since it reserves more information.
The computation and memory required for our networks and the PN-VLAD baseline are shown in Tab. \ref{tabexptime}. For our best results (FN-SF-VLAD), We have a 13.81\% increase in retrieval results (at top 1\%) at the cost of an average of 10.49ms added to per frame.

\begin{table}[!h]
\small
	\renewcommand{\arraystretch}{1.0}
	\caption{Indoor datasets evaluation results (Ave recall @ 1\%).}
\label{tabindoor}
	\centering
	\begin{tabular}{cccc}
		\hline
		\hline
		  & U.S. & R.A. & B.D. \\
		\hline
		PN-VLAD baseline  & 72.63 & 60.27 & 65.30\\
		PN-VLAD refine   & 90.10 & \bf{93.07} & 86.49\\
		FN-SF-VLAD (our)   & \bf{96.00}  & 90.46 & \bf{89.14}\\
		\hline
        \hline
	\end{tabular}
	\label{indoorresult}
\end{table}

Similar to \cite{Pointnetvlad}, we also test our network in the Indoor Dataset \cite{Pointnetvlad},
as shown in Tab.\ref{indoorresult}. Please note that we only train our netowrk on the Oxford Robotcar dataset and directly test it on the three indoor datasets, however, PointNetVLAD-refine results are obtained by training the network both on the Oxford dataset and the indoor datasets.

\subsection{Ablation Studies}\label{secablation}

\noindent {\bf Different Local Features}: We test our LPD-Net with different local features, where $xyz-$ represent the coordinates of each point, $F_{2D}$ and $F_{3D}$ are defined in Section \ref{seclocalfeature}, $FN$ represents the feature network with the proposed ten local features. In $full$, we add four $F_{3D}$ features (Planarity, Scattering, Anisotropy and Sum of eigenvalues \cite{Handfeature}) in addition to the proposed ten local features, namely, a total of 14 local features are considered. Tab. \ref{tabexplf} shows that $F_{2D}$ features have larger contributions than $F_{3D}$ features, and additional features do not contribute to improve the network accuracy since some of the features are linearly related.

\begin{table}[!h]\small
\renewcommand{\arraystretch}{1.0}
\caption{Ablation studies of different local features.}
\label{tabexplf}
\centering
\begin{tabular}{c|c|c}
\hline
\hline
 & Ave recall @1\% & Ave recall @1\\
\hline
xyz-SF-VLAD & $84.74$ & $69.75$\\
FN(non-$F_{2D}$)-SF-VLAD & $90.76$ & $76.94$\\
FN(non-$F_{3D}$)-SF-VLAD & $91.23$ & $79.11$\\
FN-SF-VLAD & $\bf{94.92}$ & $\bf{86.28}$\\
FN(full)-SF-VLAD & $92.03$ & $81.45$\\
\hline
\hline
\end{tabular}
\vspace{-0.3cm}
\end{table}

\begin{table}[!h]\small
	\renewcommand{\arraystretch}{1.0}
	\caption{Ablation studies of different feature neighbor relations.}
	\label{tabexposp}
	\centering
	\begin{tabular}{c|c|c}
		\hline
		\hline
		& Ave recall @1\% & Ave recall @1\\
		\hline
		xyz-Series-VLAD & $83.22$ & $66.01$\\
		xyz-Parallel-VLAD & $84.74$ & $69.75$\\
		FN-Original-VLAD (O) & $91.53$ & $80.29$\\
		FN-Series-VLAD (S) & $92.60$ & $81.09$\\
		FN-Parallel-VLAD (P) & $\bf{94.92}$ & $\bf{86.28}$\\
		\hline
		\hline
	\end{tabular}
\vspace{-0.2cm}
\end{table}

\begin{figure}[!h]
	\centering
	\includegraphics[width=0.8\columnwidth]{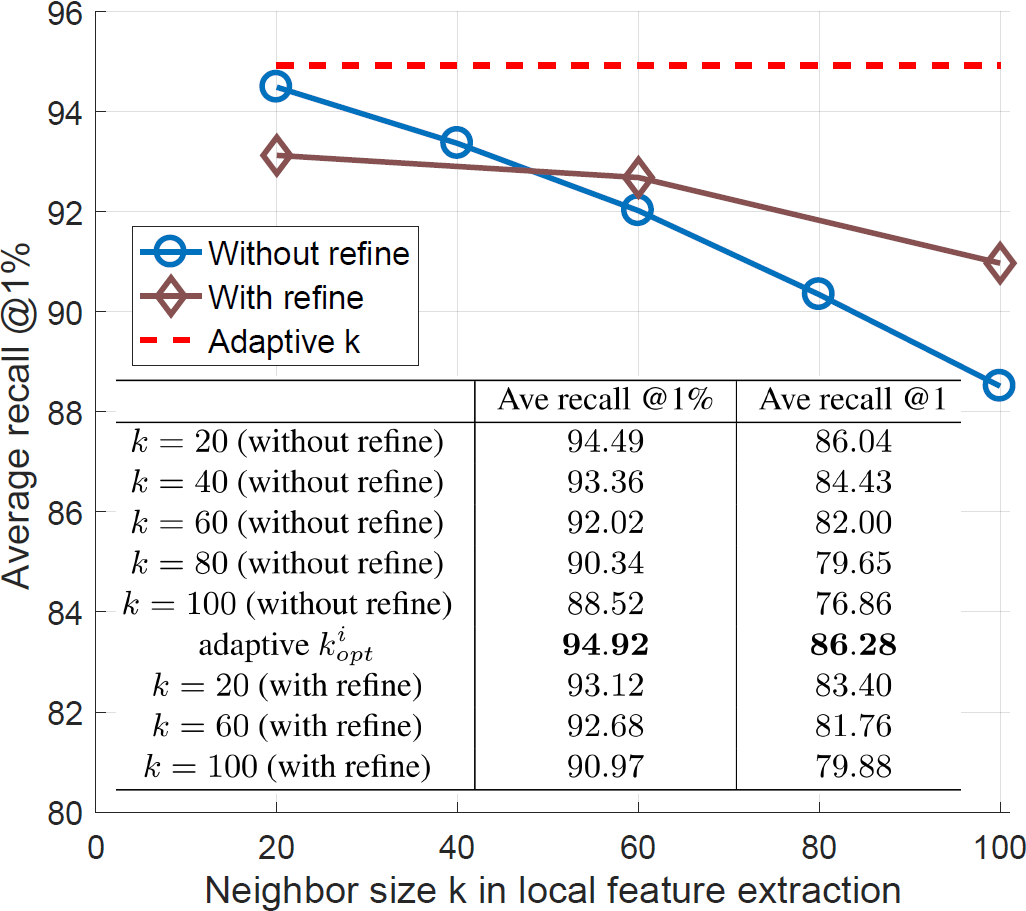}
	\caption{Ablation studies of different neighbor size $k$ in the local feature extraction.}
	\label{figexpk}
\end{figure}

\begin{table}[!h]\small
	\renewcommand{\arraystretch}{1.0}
	\caption{Ablation studies of different feature dimension $D$ and the number of visual words $K$ in NetVLAD.}
	\label{tabexpkd}
	\centering
	\begin{tabular}{c|c|c}
		\hline
		\hline
		& Ave recall @1\% & Ave recall @1\\
		\hline
		$D_{256}K_{32}$ & $93.91$ & $85.02$\\
		$D_{256}K_{64}$ & $\bf{94.92}$ & $\bf{86.28}$\\
		$D_{256}K_{128}$ & $92.47$ & $82.08$\\
		$D_{512}K_{32}$ & $92.92$ & $83.01$\\
		$D_{512}K_{64}$ & $94.66$ & $85.80$\\
		$D_{512}K_{128}$ & $93.58$ & $84.25$\\
		\hline
		\hline
	\end{tabular}
\end{table}

\begin{figure}[!h]
	\centering
	\includegraphics[width=0.98\columnwidth]{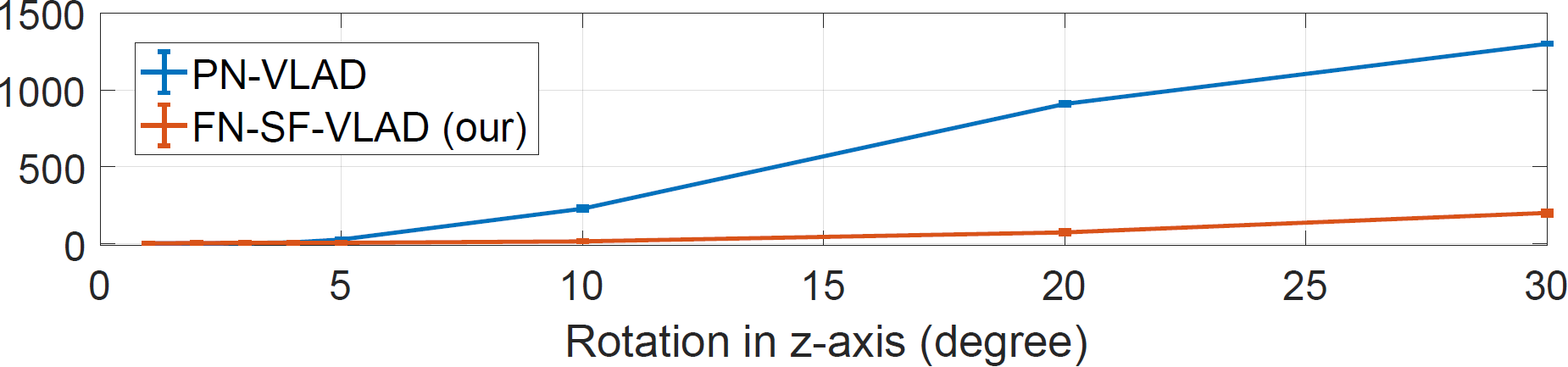}
	\caption{The number of place recognition mistakes in the robustness test.}
	\label{expfigrub}
	\vspace{-0.3cm}
\end{figure}

\begin{figure*}[!t]
	\centering
	\subfigure[]{\includegraphics[width=0.45\columnwidth]{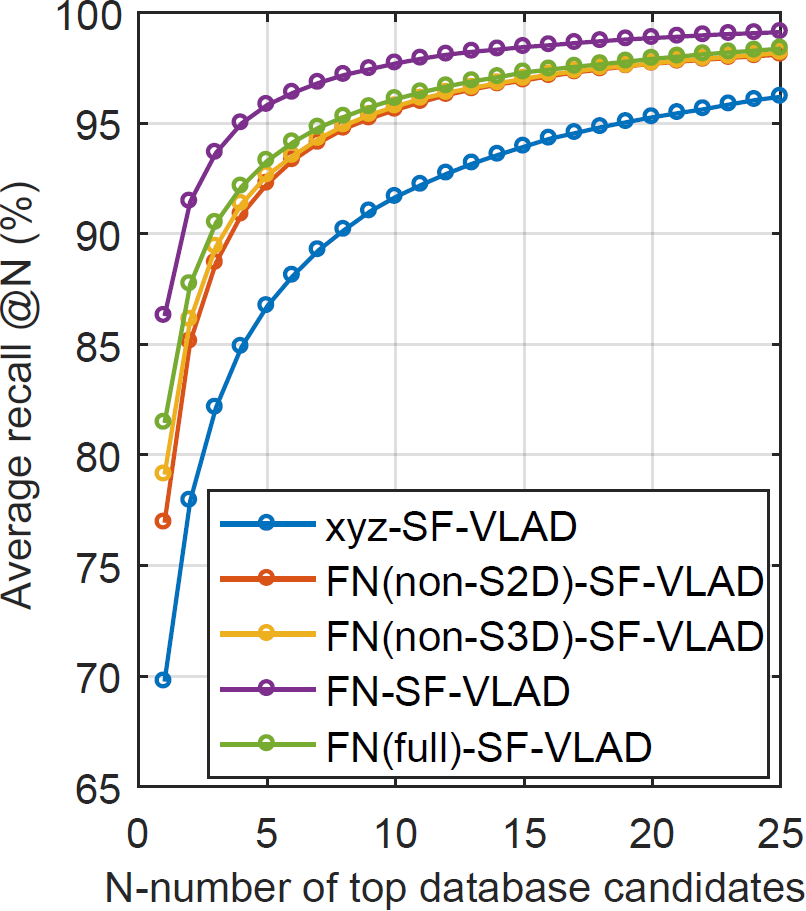} \label{figs2:1}}
	\subfigure[]{\includegraphics[width=0.45\columnwidth]{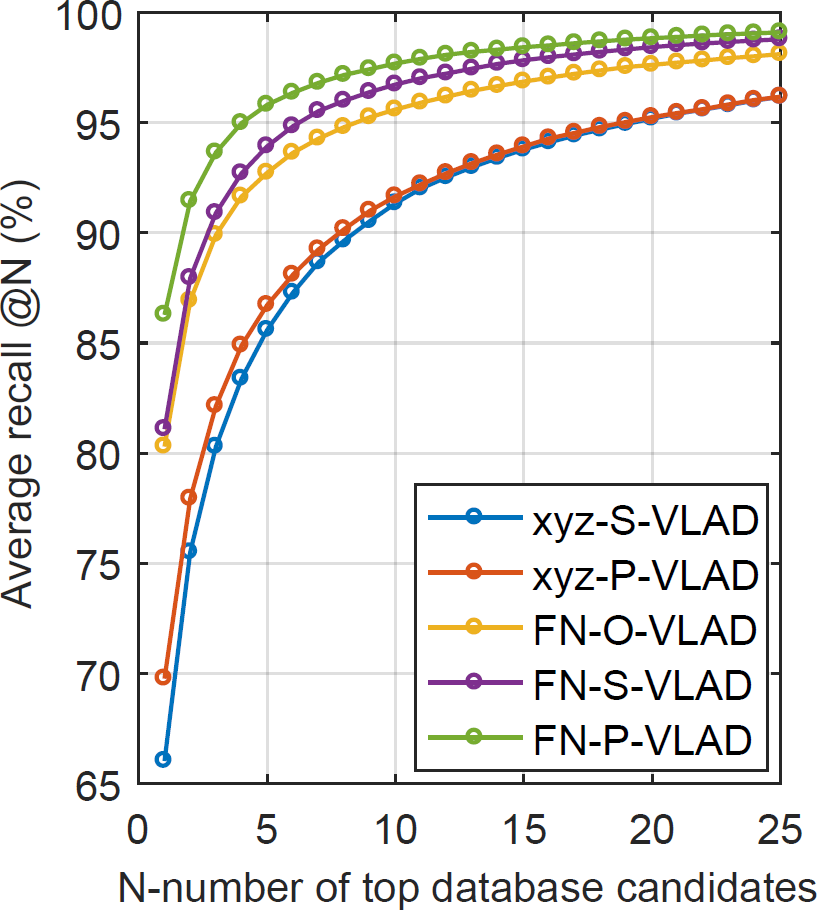} \label{figs2:2}}
	\subfigure[]{\includegraphics[width=0.45\columnwidth]{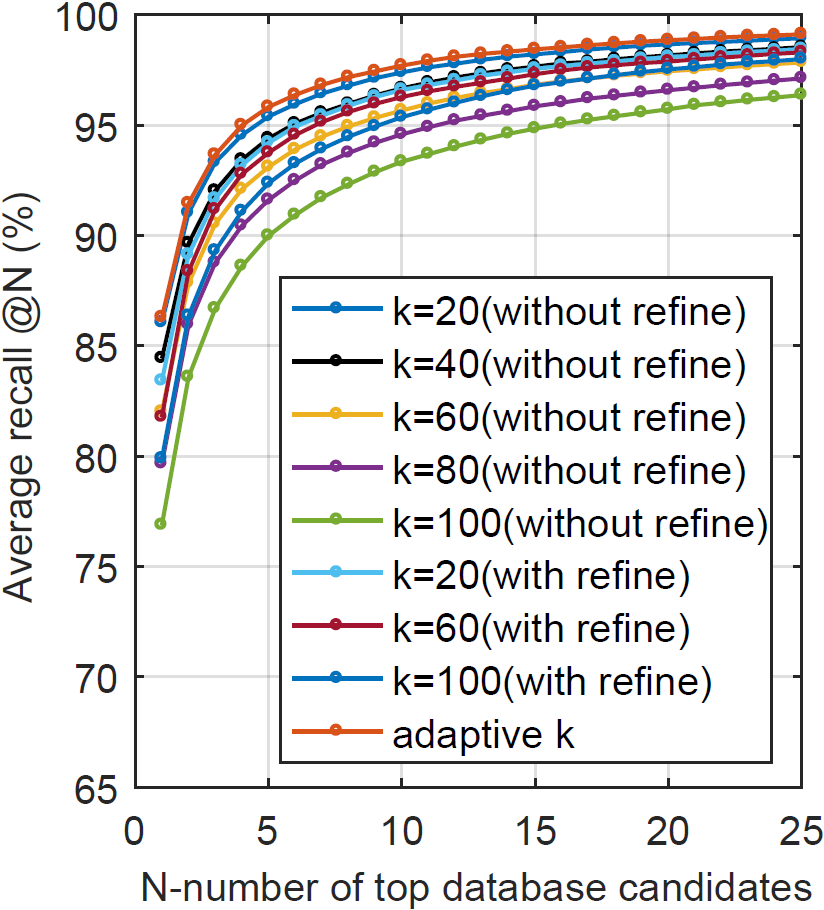} \label{figs2:3}}
	\subfigure[]{\includegraphics[width=0.45\columnwidth]{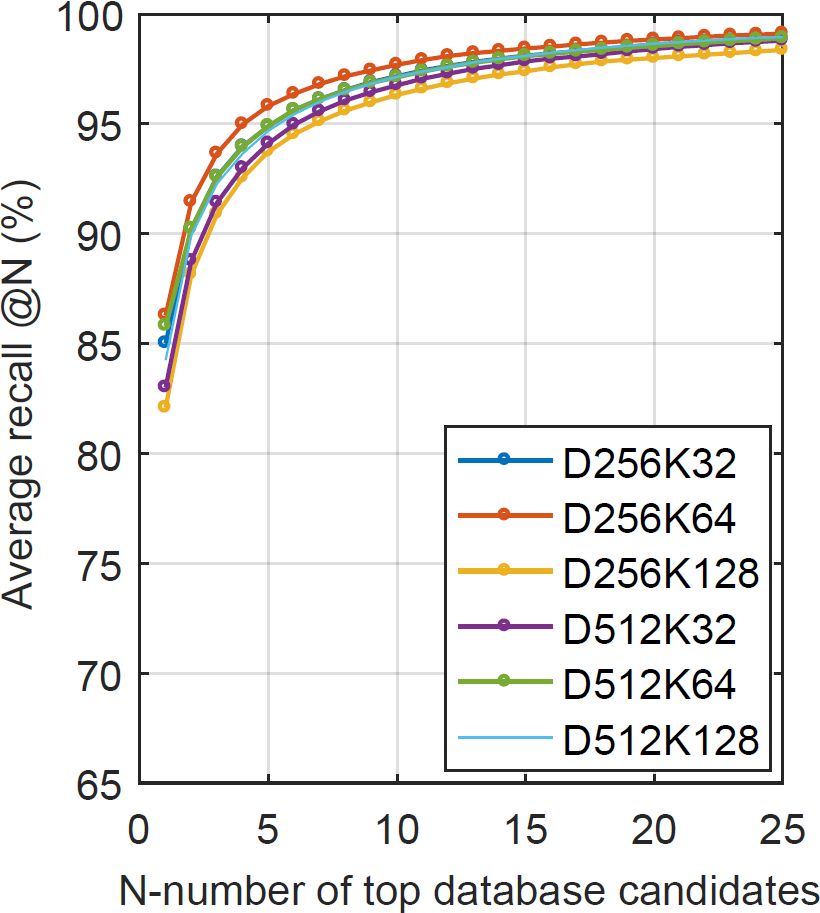} \label{figs2:3}}
	\caption{Ablation study results: (a).Different local features. (b).Different feature neighbor relations. (c).Different neighbor size $k$ in the local feature extraction.(d).Different feature dimension $D$ and the number of visual words $K$ in NetVLAD.}
	\label{expfigd}
\end{figure*}

\begin{table*}[!t]
	\myfont
	\renewcommand{\arraystretch}{0.70}
	\renewcommand\tabcolsep{5pt}
	\caption{Comparisons with vision-based methods (Ave recall @1 with different GPS location bounds: 3m/5m/10m/15m).}
	\centering
	\begin{tabular}{c|c|c|c|c|c|c|c}
		\hline
		& dawn & dusk & overcast$\_$summer &overcast$\_$winter & night-rain & sun &night\\
		\hline
		Our LPD-Net & 65.1$/$79.7$/$86.5$/$88.4 & 64.7$/$79.9$/$87.3$/$89.8 & 63.5$/$79.7$/$85.3$/$86.8 &45.6$/$73.8$/$79.2$/$81.0 &20.1$/$32.8$/$40.6$/$44.6 &74.1$/$82.3$/$87.8$/$89.4  &63.2$/$77.3$/$83.1$/$84.5\\
		HF-Net \cite{hfnet} & 45.3$/$71.2$/$81.0$/$84.7 & 54.1$/$85.8$/$92.6$/$93.9 & 55.5$/$78.8$/$83.2$/$84.7 &31.3$/$75.4$/$86.9$/$89.5 &2.7$/$6.6$/$10.5$/$11.4 &54.6$/$68.3$/$75.7$/$81.7 &2.1$/$3.9$/$7.1$/$7.3 \\ 
		NV \cite{netvlad} & 50.9$/$80.1$/$85.5$/$88.4 & 54.1$/$88.6$/$96.2$/$97.7 & 68.9$/$92.2$/$95.2$/$96.8 &29.7$/$81.0$/$94.9$/$96.7 &5.7$/$14.3$/$19.5$/$22.3 &70.0$/$82.4$/$87.6$/$89.3 &9.4$/$17.1$/$23.7$/$26.9  \\
		NV+SP \cite{hfnet}& 43.7$/$67.7$/$82.2$/$88.6 & 45.0$/$63.4$/$86.5$/$92.6 & 48.8$/$68.7$/$84.9$/$92.7 &27.2$/$60.0$/$86.7$/$93.8 &9.3$/$18.6$/$25.0$/$28.4 &48.0$/$64.3$/$84.8$/$92.4 &11.2$/$19.2$/$29.0$/$33.6  \\
		\hline
	\end{tabular}
	\label{comp}
\end{table*}

\noindent {\bf Different Feature Neighbor Relations}: We test our LPD-Net with different feature neighbor relations shown in Fig. \ref{fig:fnr}. Tab. \ref{tabexposp} shows that $P$ is better than $O$ and $S$, which implies that only utilizing the feature relations in the transformed feature space and remaining the original feature vectors can achieve the best result. Please note that in PointNet and PointNetVLAD, they use the $S$ relation.

\noindent {\bf Different Neighbor Size $k$ in the Local Feature Extraction}: Fig. \ref{figexpk} shows that, in the case of constant $k$, the accuracy decreases with the size of $k$. With refinements (retrain the network with the fixed $k$), the accuracy is still lower than that of the proposed adaptive approach ($k_{opt}^i$).

\noindent {\bf Different $K$ and $D$ in NetVLAD}: NetVLAD has two unique hyper-parameters: the feature dimension $D$ and the number of visual words $K$ \cite{netvlad,Pointnetvlad}. Tab. \ref{tabexpkd} shows that the values of $K$ and $D$ should be matched in order to achieve a good accuracy. We use $K = 64$ and $D = 256$ in this paper.

All the above ablation studies are conducted on the robotcar dataset. The detailed results are shown in Fig. \ref{expfigd}.

\noindent {\bf Robustness Test}: We rotate the input point cloud and add $10\%$ random white noise to validate the robustness of our LPD-Net. The results are shown in Fig. \ref{expfigrub}, more details can be found in our supplementary materials. 

\subsection{Comparison with image-based methods}

\begin{figure}[!h]
	\centering
	\includegraphics[width=1\columnwidth]{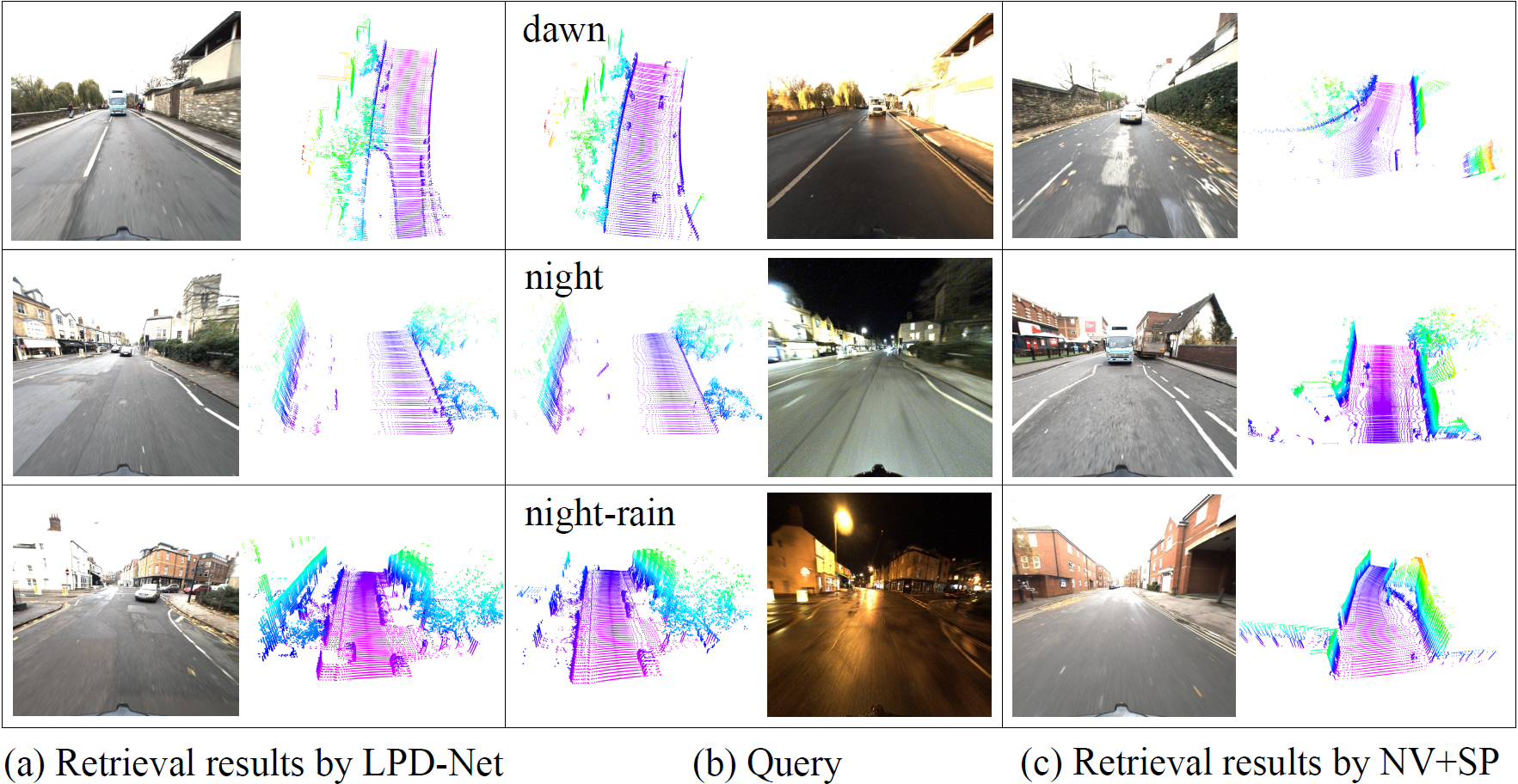}
	\caption{Examples of the retrieval results of our LPD-Net and the image-based solution NV+SP \cite{hfnet}. The middle column shows the query images and point clouds, the left column shows the retrieved point clouds by LPD-Net and their corresponding images, the right column shows the retrieved images by NV+SP and their corresponding point clouds.}
	\label{expfigcomp}
	\vspace{-0.5cm}
\end{figure}

To further investigate the advantages of our LPD-Net, the preliminary comparison results with the state-of-the-art image-based solutions are shown in Tab. \ref{comp}, where NV is a pure NetVLAD method, HF-Net and NV+SP are proposed in \cite{hfnet}. This comparison is conducted on the Robotcar Seasons dataset \cite{benchmark}, and we generate the corresponding point clouds by using the original data from the Robotcar dataset.
We can observe that, in the most of the cases, our point cloud-based method shows strong performance on par or even better than image-based methods. A special case lies in the \textit{night-rain} scene, since the point cloud data used here is reconstructed using a single-line LiDAR and visual odometry (VO), the inaccuracy of VO causes the point cloud to be distorted, hence resulting in a reduced result. However, we can still observe that our method significantly outperforms other approachs in the \textit{night-rain} case. Fig. \ref{expfigcomp} shows three examples in different cases. In these examples, the image-based solution obtains the unsuccessfully retrieved images, due to the bad weather and light conditions. However, our LPD-Net obtains the correct results.

Please noted that the presented work at this stage only focuses on the point cloud-based place recognition, however, the above image-based solutions are proposed for the pose estimation task, so the above comparisons are not rigorous. In the future, we will improve our LPD-Net in order to solve the pose estimation problem. 

\section{Conclusion}
In this paper, we present the LPD-Net that solves the large-scale point cloud-based retrieval problem so that the reliable place recognition can be successfully performed. 
Experimental results on benchmark datasets validate that our LPD-Net is much better than PointNetVLAD and reaches the state-of-the-art. What's more, comparison results with image-based solutions validate the robustness of our LPD-Net under different weather and light conditions.


\onecolumn 
\section{Supplementary Material}

\subsection{Detailed differences between our network and existing networks}\label{S2}

Comparison of the proposed network with existing methods is shown in Table \ref{suptabreview}.
PointNet \cite{pointnet} provides a simple and efficient point cloud feature learning framework that directly uses the raw point cloud data as the input of the network, but failed to capture fine-grained patterns of the point cloud due to the ignored local features of points. PointNet++ \cite{pointnet++} considers the hierarchical feature learning, but it still only operates each point independently during the local feature learning process, which ignores the relationship between points, leading to the missing of local feature.
Then, KCNet \cite{kcnet} and DGCNN \cite{dgcnn} have been designed based on PointNet. With the kNN-based local feature aggregating in the feature space, the state-of-the-art classification and segmentation results are obtained on small-scale object point cloud dataset. Moreover, RWTH-Net \cite{rwthnet} achieved fine-grained segmentation results on the large-scale outdoor dataset based on local feature aggregating and clustering in Feature space and Cartesian space, respectively.
However, the above networks all focus on capturing the similarity of point clouds, and it is hard to generate effective global descriptors of the point cloud in large scenes.
On the other hand, PointNetVLAD \cite{Pointnetvlad} can learn the global descriptor of the point cloud in a large scene, but it only performs feature aggregation in the feature space, ignoring the distribution of features in Cartesian space, which makes it difficult to generalize the learned features.

\begin{table}[!h]
	\renewcommand{\arraystretch}{1.3}
	\caption{Where our work fits into the literature.}
    \label{suptabreview}
	\centering
	\begin{tabular}{|c|c|c|c|c|c|}
		\hline
		 & \tabincell{c}{Local\\features} & \tabincell{c}{Feature space\\aggregation} & \tabincell{c}{Cartesian space\\aggregation} & \tabincell{c}{Feature\\ distribution} & \tabincell{c}{Large-scale\\ scene} \\
		\hline
		PointNet \cite{pointnet} &  & & & &  \\
		\hline
		PointNet++ \cite{pointnet++} &  &   & $\checkmark$  &  &  \\
		\hline
		PointNetVLAD \cite{Pointnetvlad} &  & $\checkmark$ &  &  & $\checkmark$ \\
		\hline
		DGCNN \cite{dgcnn} &  & $\checkmark$ &  &  &  \\
		\hline
		KCNet \cite{kcnet} &  & $\checkmark$ &  &  &  \\
		\hline
		RWTH-Net \cite{rwthnet} & &$\checkmark$ & $\checkmark$ &  & $\checkmark$  \\
		\hline
		Our LPD-Net & $\checkmark$ & $\checkmark$  & $\checkmark$ & $\checkmark$ & $\checkmark$  \\
		\hline
	\end{tabular}
\end{table}

\subsection{Detailed environment analysis results}\label{S5}

\begin{figure}[!h]
	\centering
	\includegraphics[width=0.9\columnwidth]{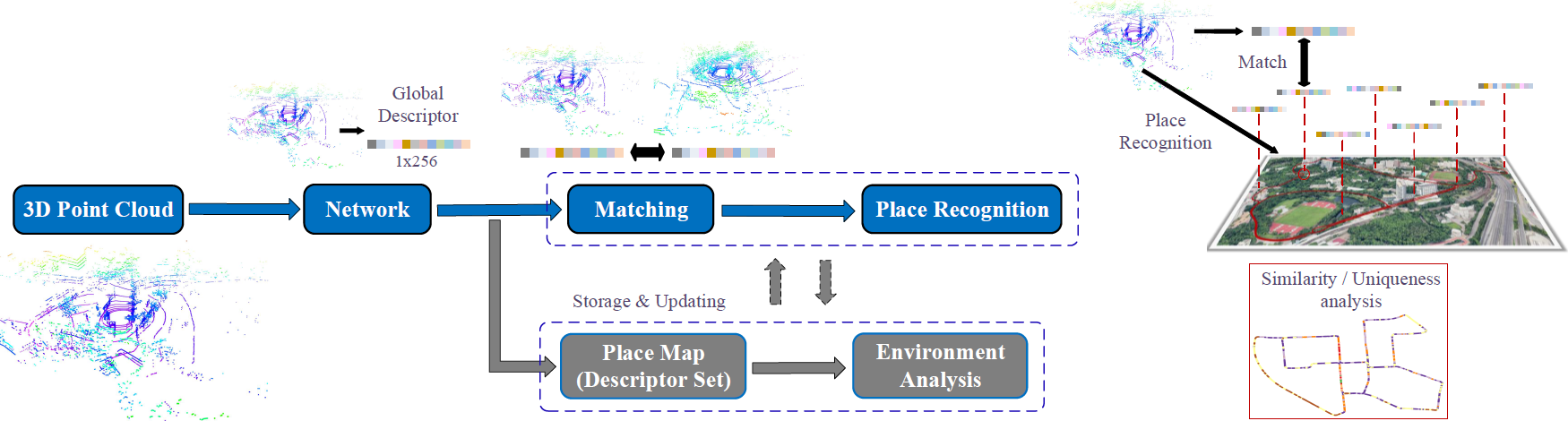}
	\caption{System structure.}
	\label{supfig:sysstructure}
\end{figure}

\begin{figure}[!h]
	\centering
	\includegraphics[width=0.5\columnwidth]{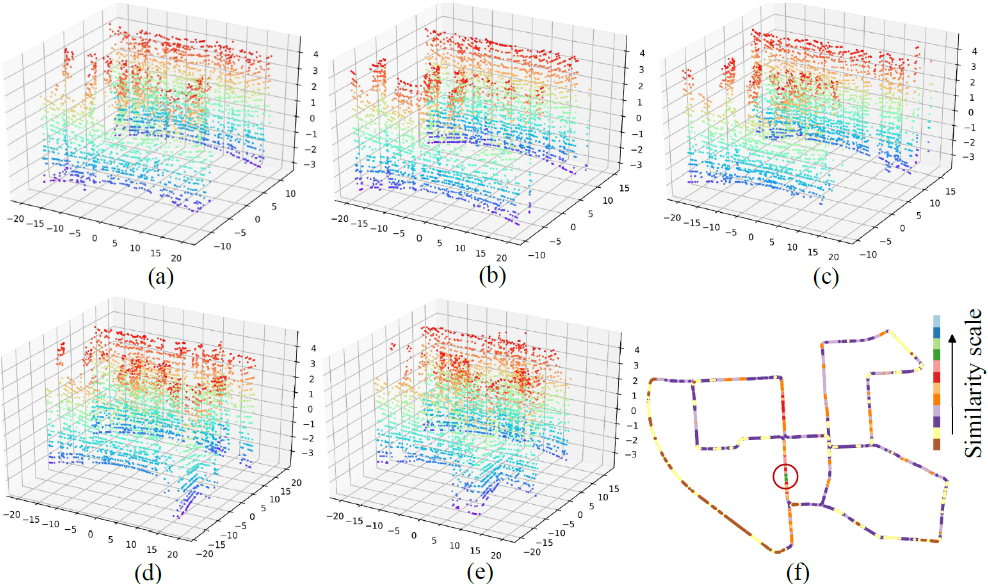}
	\caption{Similarity evaluation. (a)-(e) show a sequence of point clouds with near locations (correspond to the places within the red circle in f). (f) shows the similarity between the point cloud in (b) and all the other point clouds in the whole environment.}
	\label{supfigsimilar}
	\vspace{-0.3cm}
\end{figure}

\begin{figure}[!h]
	\centering
	\includegraphics[width=0.5\columnwidth]{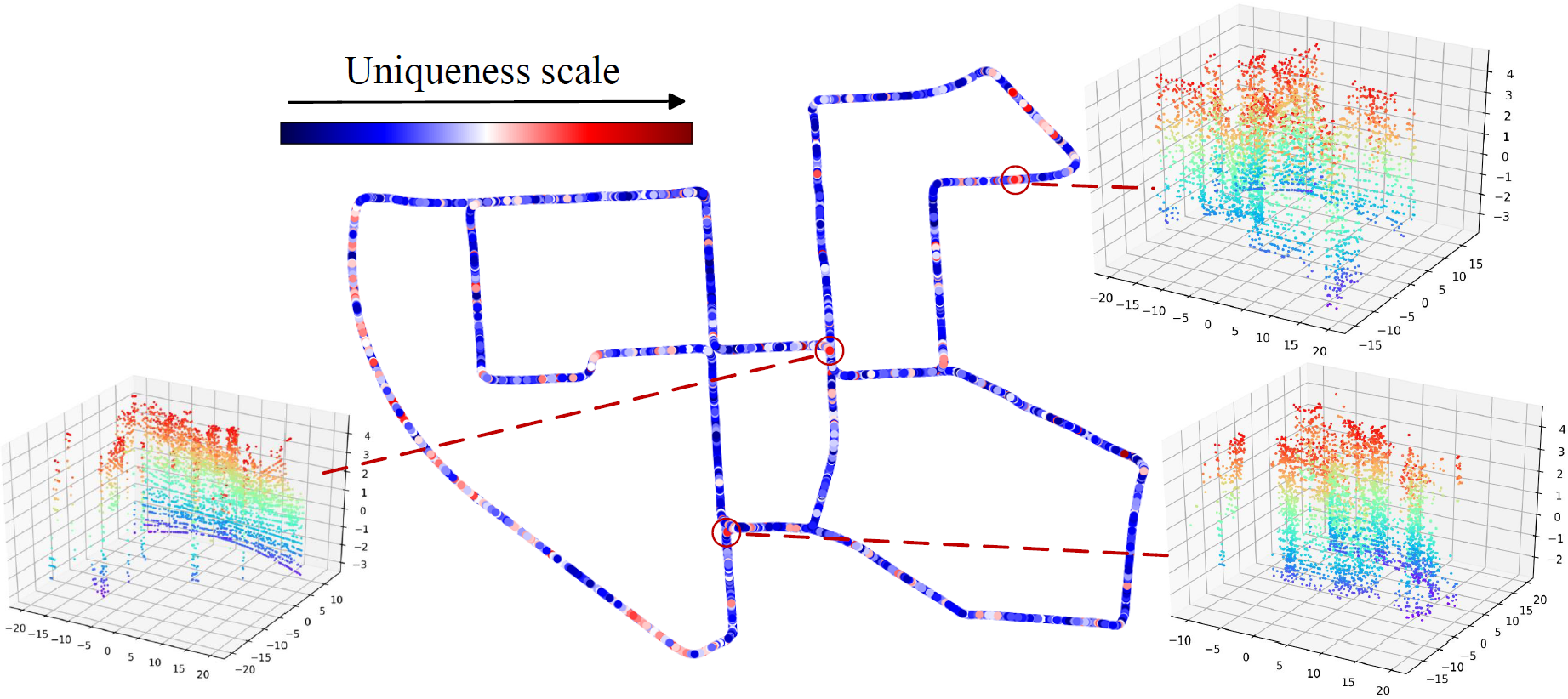}
	\caption{Uniqueness evaluation. The corresponding point clouds of three places with high uniqueness are shown.}
	\label{supfigunique}
	\vspace{-0.3cm}
\end{figure}

Our LPD-Net extracts discriminative and generalizable global descriptors of an input point cloud, so we can resolve the large-scale place recognition and environment analysis problems. 
The system structure is shown in Fig. \ref{supfig:sysstructure}. The original 3D Lidar point cloud is used as system input directly. We use LPD-Net to extract a global descriptor which will be stored in a descriptor set.
On the one hand, when a new input point cloud is obtained, we match its descriptor with those in the descriptor set to detect that whether the new scene corresponds to a previously visited place, if so, a loop closure detection accomplished. 
On the other hand, we analyze the environment by investigating the statistical characteristics of the global descriptors of the already visited places to evaluate the similarity/uniqueness of each place. 
The similarity between two point clouds is calculated from the distance between two corresponding global descriptors. Then for a given point cloud, we can calculate the similarity index of this point and plot a similarity map as shown in Fig. \ref{supfigsimilar} (in KITTI dataset). With normalization, the sum of the similarity with all the other point clouds in the whole environment can be used to evaluate the uniqueness of the given point cloud (the given place). Fig. \ref{supfigunique} shows the uniqueness evaluation results of the whole environment in KITTI dataset.

\begin{figure}[h]
	\centering
	\includegraphics[width=0.75\columnwidth]{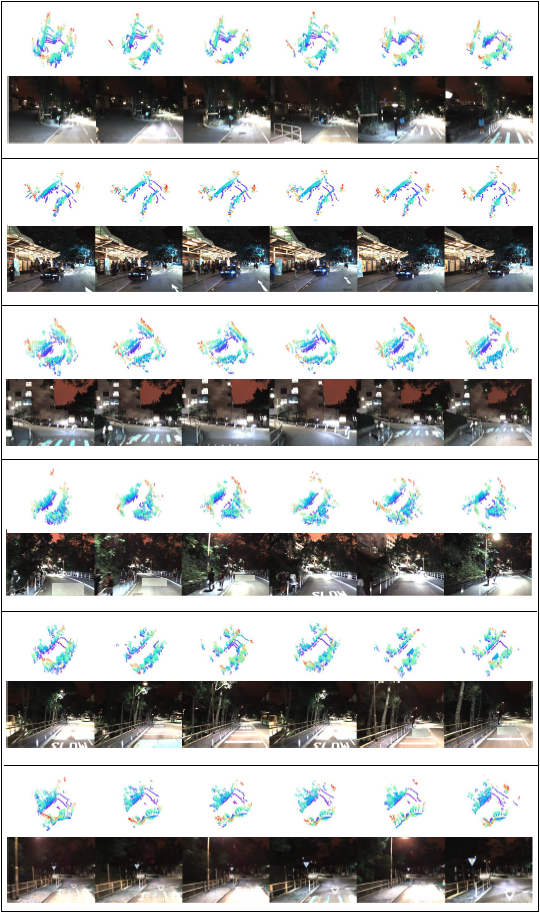}
	\caption{Examples of the place clustering results: the clustered places which are also near with each other in the geographical location.} 
	\label{supfigmat1}
\end{figure}

\begin{figure}[h]
	\centering
	\includegraphics[width=0.75\columnwidth]{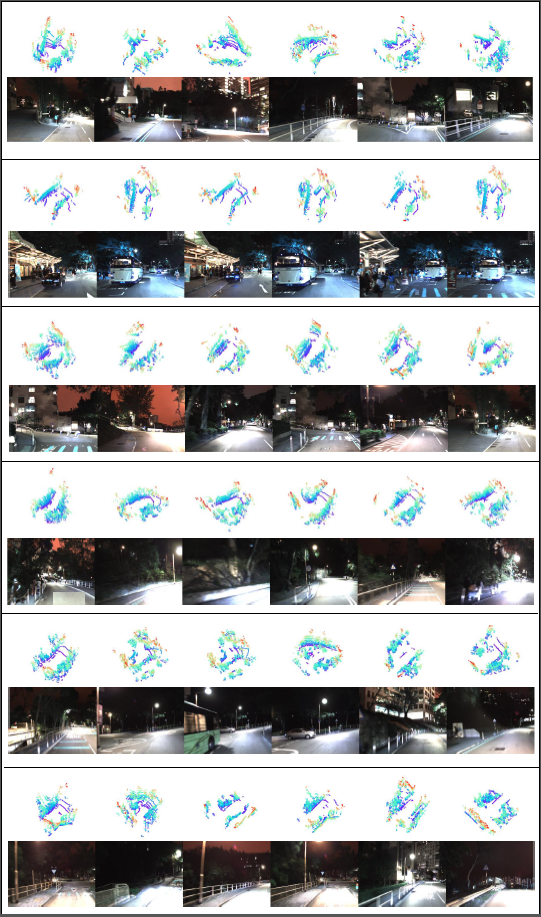}
	\caption{Examples of the place clustering results: the clustered places which are far from each other in the geographical location.}
	\label{supfigmat2}
\end{figure}

We also cluster the global descriptors generated by our LPD-Net to see the places in the same cluster. We collect the point cloud data with a VLP-16 LADAR in our university campus. In Fig. \ref{supfigmat1}, we show the clustered places which are also near with each other in the geographical location, and in Fig. \ref{supfigmat2}, we show the clustered places which are far from each other in the geographical location.

\subsection{Detailed robustness test results}\label{S6}

\begin{figure}[h]
	\centering
	\includegraphics[width=0.65\columnwidth]{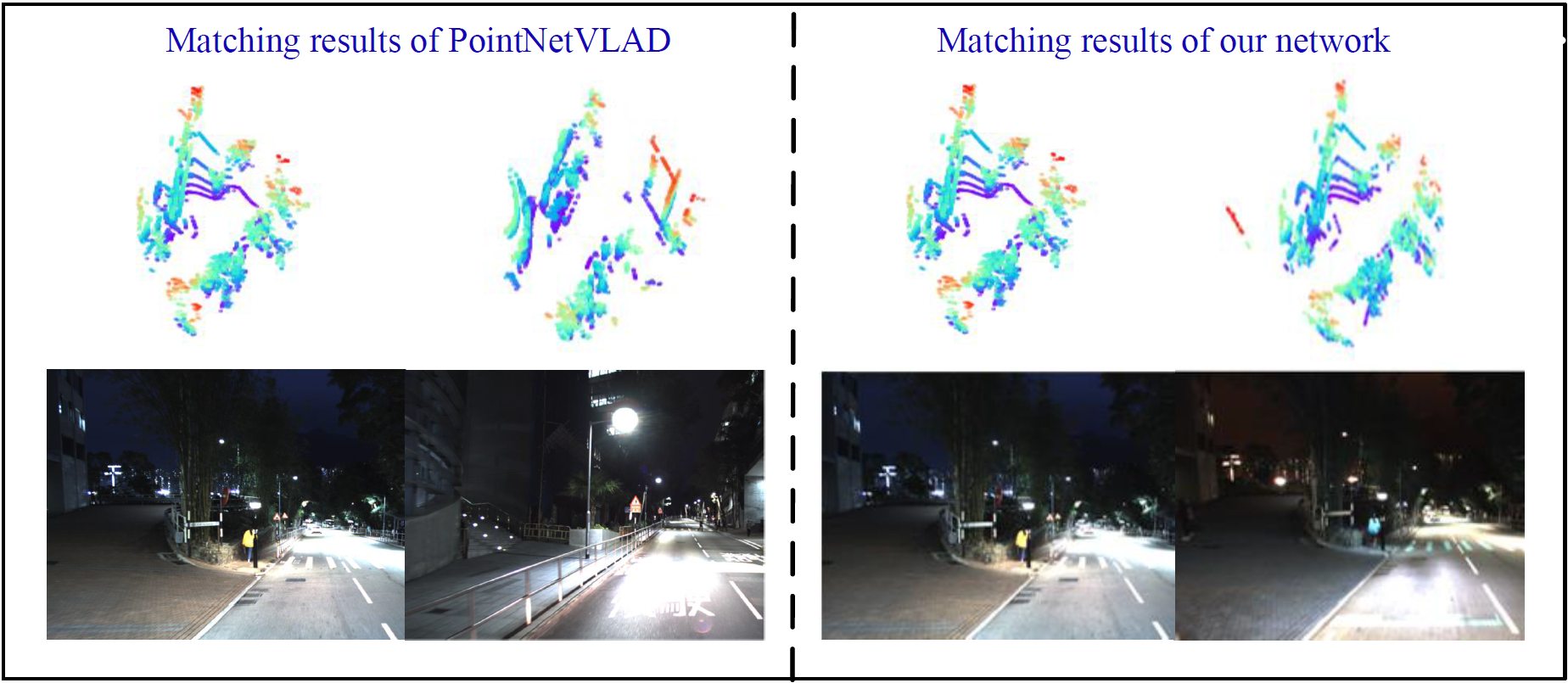}
	\caption{An Example of the failure cases in PointNetVLAD in the robustness tests. In this case, we rotate the input point cloud with 10 degrees and add 10\% random white noise. Our network recognizes the place correctly, however, PointNetVLAD retrieves a wrong place.}
	\label{supfigrob}
\end{figure}

The detailed results of the robustness test presented in the main paper are shown in Tab. \ref{suptabrota} and Fig. \ref{supfigrob}. In Tab. \ref{suptabrota}, we show the average (max) number of place recognition mistakes under different degrees of input point cloud rotations. We consider 1580 places in the university campus dataset and also add 10\% white noise in the tests. Both results from our network and PointNetVLAD are shown. Fig. \ref{supfigrob} shows an example of the failure cases in PointNetVLAD in the robustness tests.

\begin{table}[h]
	\renewcommand{\arraystretch}{1.3}
	\caption{Results in the robustness test: with different rotation angles (degree).}
    \label{suptabrota}
	\centering
	\begin{tabular}{|c|c|c|c|c|c|c|c|c|}
		\hline
		 & 1 & 2 & 3 & 4 & 5 & 10 & 20 & 30 \\
		\hline
		PointNetVLAD & 4.5(6) & 5.25(7) & 7.75(13) & 10.75(12) & 29.25(31) & 229.25(232) & 908.75(918) & 1298.75(1301) \\
		\hline
		Our LPD-Net & 4.75(6) & 6(8) & 6.5(8) & 7.25(9) & 8(10) & 17.25(21) & 75.5(87) & 202(217) \\
		\hline
	\end{tabular}
    \begin{tablenotes}
		\item[1] A(B): A is the average number across eight repeated experiments in each case, B is the max number in the eight experiments.
	\end{tablenotes}
\end{table}

\end{document}